\documentclass[journal]{IEEEtran}
\usepackage{amsmath,amsfonts}
\usepackage{algorithmic}
\usepackage{algorithm}
\usepackage{array}
\usepackage{multirow}
\usepackage[caption=false,font=normalsize,labelfont=sf,textfont=sf]{subfig}
\usepackage{textcomp}
\usepackage{stfloats}
\usepackage{url}
\usepackage{verbatim}
\usepackage{float}
\usepackage{graphicx}
\usepackage{cite}
\usepackage{gensymb}
\usepackage[table,xcdraw]{xcolor}
\begin{document}
\title{A 360\degree{} Multi-camera System for Blue Emergency Light Detection Using Color Attention RT-DETR and the ABLDataset}

\author{%
Francisco Vacalebri-Lloret, %
Lucas Banchero, %
Jose J. Lopez %
and Jos\'e M. Mossi%
\thanks{This project was funded by the project AEOLIAN:OMAP (TED2021-131003B-C22) in PTRS by the Spanish Government.}
\thanks{F. Vacalebri-Lloret, L. Banchero, J. J. Lopez and J. M. Mossi are with the Institute of Telecommunications and Multimedia Applications, Universitat Polit\`ecnica de Val\`encia, 46022 Val\`encia, Spain. 
(e-mail: \{fvacllo, lbanmar\}@iteam.upv.es; \{jmmossi, jjlopez\}@dcom.upv.es).}
\thanks{Corresponding author: J.M. Mossi (e-mail: jmmossi@dcom.upv.es).}
\thanks{\textit{This work has been submitted to IEEE for possible publication.}}
}

\maketitle

\begin{abstract}
This study presents an advanced system for detecting blue lights on emergency vehicles, developed using ABLDataset, a curated dataset that includes images of European emergency vehicles under various climatic and geographic conditions. The system employs a configuration of four fisheye cameras, each with a 180\degree{} horizontal field of view, mounted on the sides of the vehicle. A calibration process enables the azimuthal localization of the detections. Additionally, a comparative analysis of major deep neural network algorithms was conducted, including YOLO (v5, v8, and v10), RetinaNet, Faster R-CNN, and RT-DETR. RT-DETR was selected as the base model and enhanced through the incorporation of a color attention block, achieving an accuracy of 94.7\% and a recall of 94.1\% on the test set, with field test detections reaching up to 70 meters. Furthermore, the system estimates the approach angle of the emergency vehicle relative to the center of the car using geometric transformations. Designed for integration into a multimodal system that combines visual and acoustic data, this system has demonstrated high efficiency, offering a promising approach to enhancing Advanced Driver Assistance Systems (ADAS) and road safety.
\end{abstract}

\begin{IEEEkeywords}
Real-time object detection, Transformer-based vision, Color attention, Blue Light Detection, Emergency Vehicle Location, Fisheye Multicamera system, 360\degree{} surround view, Advanced Driver Assistance Systems (ADAS), Automotive edge computing.
\end{IEEEkeywords}

\section{Introduction} \label{sec:intro}
\IEEEPARstart{T}{he} automatic detection of emergency vehicles is a critical component in the development of intelligent transportation systems and driver assistance technologies. In emergency situations, the rapid and accurate identification of these vehicles is crucial for improving road safety, reducing response times, and minimizing risks for both drivers and emergency teams \cite{Zohaib2024EnhancingFusion}.

In the field of road safety, blue emergency lights have been widely adopted in Europe and globally due to their high visibility and ability to alert drivers to critical situations. According to a study by Bullough \emph{et al.} \cite{Bullough2019ImpactsGlare}, blue lights generate a greater perception of brightness and significantly contribute to drivers’ ability to identify priority vehicles, especially in low-light conditions. The same study indicates that although blue lights are more prone to causing glare, they enhance driver response in emergency situations compared to lights of other colors.

UNECE Regulation No. 48 \cite{UNECE2016RegulationUNECE} clearly specifies the use of lighting devices in vehicles, including those intended for emergency signaling functions. The regulation establishes the permitted colors for vehicle-emitted lights, including the use of blue light for specific functions. This highlights the widespread adoption of these lights in Europe, supported by harmonized regulations aimed at ensuring their uniform implementation across all member countries. Additionally, records indicate that blue lights on priority emergency vehicles are also extensively used in other regions, such as North America and Asia, where they serve similar functions in emergency vehicles, highlighting their global importance in road safety and emergency management.

\begin{figure}[!ht]
    \centering
    \includegraphics[width=\columnwidth]{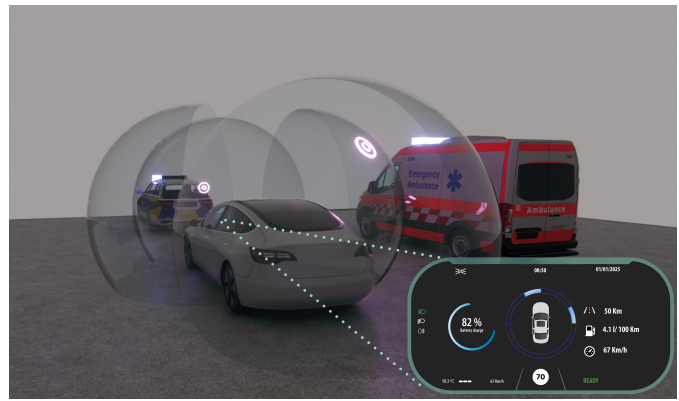}
    \caption{Illustrative example of the proposed system.}
    \label{fig:fig1}
\end{figure}

Recent studies indicate that anticipating an emergency vehicle’s (EV) movements and providing the driver with early alerts can have a positive impact on road safety. Data mining analyses of collision patterns suggest that crash severity decreases when drivers receive signals allowing them more time to respond \cite{Hossain2023DataModes,Hsiao2018PreventingIssues}. Along these lines, the development of in-cabin alert systems—capable of detecting and conveying the presence of an approaching EV—has proven effective at helping drivers shift to a safer lane more quickly and adapt their driving behavior \cite{Weibull2025DriversIntroduction}. Figure~\ref{fig:fig1} provides an example of how such an active emergency vehicle alert system could operate.

\begin{figure*}[!ht]
    \centering
    \includegraphics[width=\textwidth]{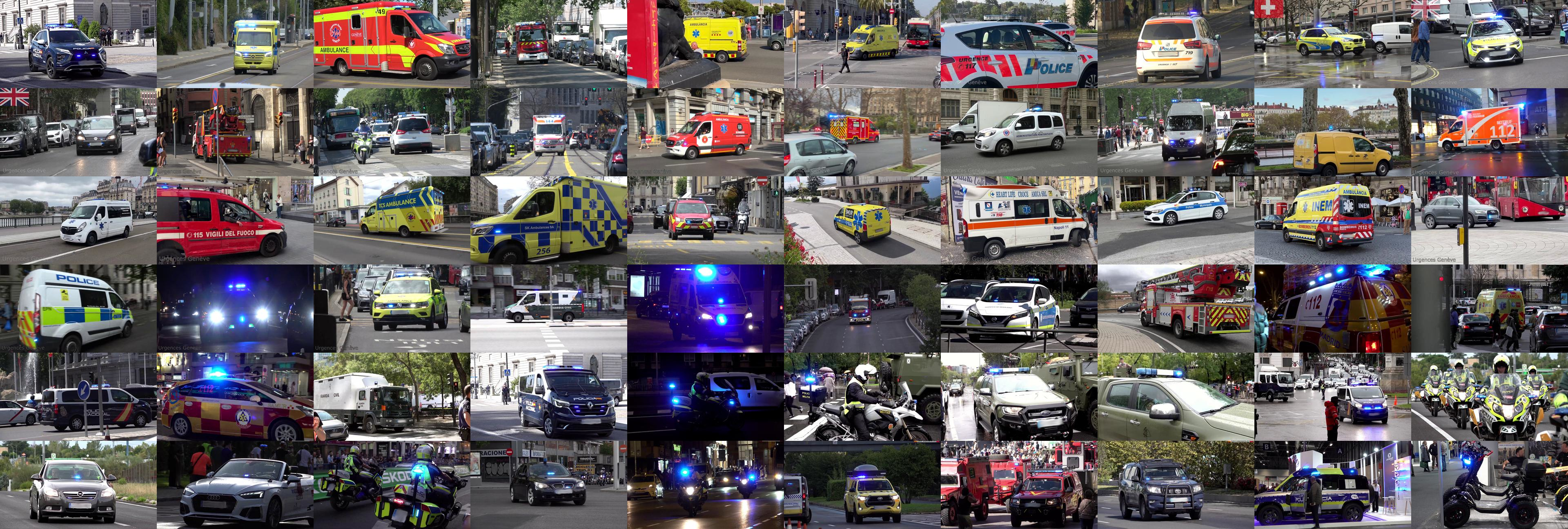}
    \caption{Sample of the variety of emergency vehicles considered in ABLDataset.}
    \label{fig:fig2}
\end{figure*}

In the context of Advanced Driver Assistance Systems (ADAS), emergency light detection emerges as a solution that enables the generalized identification of priority vehicles across a wide range of countries and contexts worldwide. Many vision-based studies on the detection of these vehicles rely on specific visual features \cite{Ashir2022ADetection,Kaushik2020LeveragingAnalysis}, whose appearance may vary significantly not only between countries but even between neighboring municipalities (Fig.~\ref{fig:fig2}).

On the other hand, numerous studies focus on detecting sirens through audio \cite{Lisov2023UsingDetection,Mittal2022AcousticModels,Parineh2023DetectingSignals,Banchero2025EnhancingSound} or on the implementation of multimodal systems \cite{Mecocci2024RTAIAED:YOLOv8,Tran2021Audio-VisionDetection,Joshi2024Multi-ModalVehicles,Zohaib2024EnhancingFusion}. However, there are situations, particularly at night, in which emergency vehicles operate without activating the siren but with emergency lights on. This makes emergency lights a constant and essential element in the operation of such vehicles, establishing them as a key feature for their detection.

Currently, vehicles with a Level~3 or higher autonomy, according to the SAE autonomy scale \cite{SAEInternationalRecommendedPractice2021TaxonomyVehicles} are equipped with advanced visual perception systems that capture images of the immediate environment to analyze the scene and make real-time decisions \cite{Badue2021Self-drivingSurvey}. The positioning of cameras varies depending on the vehicle model, with possible integration into the front grille, side mirrors, B-pillars, rear section, or even the roof to cover multiple viewing angles \cite{Yogamani2019WoodScape:Driving}. In many cases, fisheye cameras are used due to their wide field of view, allowing for an extensive observation area with a single capture device. These cameras are typically integrated into the vehicle’s body, connected via high-speed internal buses to a central processing unit or distributed computing units \cite{Hasan2022OpticalReview}. Additionally, vehicular communication networks are migrating towards higher bandwidth standards, such as automotive Ethernet, Time-Sensitive Networking (TSN) protocols, and CAN FD \cite{Peng2023ANetworking,ChehidaDouss2023State-of-the-artVulnerabilities}, to process multimodal sensor data in a synchronized and real-time manner. This infrastructure ensures the low latency and reliability necessary for safe and efficient decision-making in driver assistance systems.

In this context, ADAS has become an essential component for enhancing driver safety and experience \cite{Gonzalez-Saavedra2022SurveyVision}. Improvements in computer vision algorithm performance, along with the integration of complementary sensors (radars, LiDARs, microphones, etc.), have enabled the implementation of functionalities such as lane-keeping assistance, automatic emergency braking, traffic sign detection and recognition, and 360\degree{} vehicle environment monitoring, among many others \cite{Li2022AChina}. In this study, our objective is to explore a new type of ADAS system aimed at the early detection of emergency vehicles, representing a significant advancement in reducing response times and preventing accidents in critical situations.

The ability to achieve a 360\degree{} surround view through multiple cameras provides significant advantages, allowing for the early detection of key elements that the driver might not perceive at first glance, thereby reducing reaction time to unexpected events \cite{Kumar2021OmniDet:Driving}. Visual concentration in a single direction of traffic can lead to the omission of relevant information in other areas, increasing the risk of collisions or inadequate anticipation. Therefore, a system capable of automatically alerting the driver to the presence and direction of priority vehicles—through integrated dashboard screens or Head-Up Displays (HUDs)—enhances decision-making, contributing to greater safety and efficiency in traffic flow \cite{StojmenovaPececnik2023DesignVehicles} (Fig.~\ref{fig:fig1}). This alert could be synchronized with other vehicle systems (e.g., electronic mirror adjustments or integrated acoustic warnings), further improving the driver’s response capacity.

Within this framework, the present study introduces the research and development of a system capable of accurately detecting blue emergency lights and determining the approach angle of a priority vehicle, leveraging the multi-camera configurations that are already standard equipment in modern automobiles \cite{VishwasVenkat2023ReviewVehicles,Kumar2023Surround-ViewChallenges}. The combination of panoramic visual information with geometric calibration techniques and data fusion enables this system to provide spatial information, laying the groundwork for future advancements in vehicular intelligence and driver assistance strategies.

The evolution of Artificial Intelligence (AI) applied to computer vision has driven the design of multiple deep neural network architectures, enabling the detection and classification of objects with a high degree of accuracy and robustness \cite{Zhao2019ObjectReview,Liu2020DeepSurvey,Zou2023ObjectSurvey}. These advancements have been particularly significant in the automotive sector, where the vehicle—understood as a computational node with onboard processing resources—can differentiate, identify, and track its surrounding environment in real time \cite{Kumar2021OmniDet:Driving,Wang2023VV-YOLO:YOLOv4,Dazlee2022ObjectYOLO}. Along with the progressive miniaturization of components, increased energy efficiency, and the growing use of specialized hardware accelerators (GPUs, TPUs, ASICs, and FPGAs) \cite{Lee2023AnalysisVehicles,Lu2023VehicleChallenges}, it has become feasible to implement these AI solutions at the edge (Edge Computing), ensuring minimal latency and reducing reliance on connectivity with external infrastructures. This continuously expanding technological landscape provides a favorable environment for the deployment of advanced emergency vehicle detection systems, maximizing their utility in situations where the driver's reaction time can be critical.

For the proper development of automatic detection systems, high-quality datasets are essential. Among the most relevant datasets that include images captured using 360\degree{} multi-camera systems integrated into vehicles are WoodScape \cite{Yogamani2019WoodScape:Driving}, FisheyeDataset \cite{ADASBiro2023FisheyeDatasetDataset}, ADD \cite{Wu2023ADD:Driving} and other similar projects that do not fully meet the criteria of 360\degree{} coverage or fisheye camera use but remain relevant, such as KITTI \cite{Geiger2013VisionDataset} or Cityscapes \cite{Cordts2016TheUnderstanding}. These datasets provide a wide variety of labeled traffic scenes under diverse environmental and road conditions, with some of them also incorporating fisheye cameras. To address this gap, this paper introduces the \emph{Active Blue Light Dataset (ABLDataset)}, a specialized dataset designed specifically for the detection of active blue emergency lights in real-world traffic scenarios.

The information contained in these sources has been valuable in understanding and adjusting the characteristic distortion parameters of fisheye lenses, as well as in verifying the suitability of the four-camera arrangement in the vehicle. However, the lack of specific examples featuring emergency vehicles with active blue lights has limited the direct applicability of these datasets for training the proposed system. The absence of annotations focused on this particular element has highlighted the need for a dedicated dataset capable of capturing the variability in visual conditions, geographic settings, and the aesthetic particularities of emergency vehicles across different countries. To address this gap, in this work we have created the ABLDataset, which provides a comprehensive resource tailored specifically for this purpose.

In this regard, there is relatively scarce literature that has systematically addressed the detection of emergency vehicles using blue lights as the primary visual feature. This gap opens the door to a new line of research, which this study aims to initiate. By providing a more specialized approach, this research seeks to promote the development of methods that incorporate this critical chromatic information into the detection of priority vehicles, enhancing the capabilities of driver assistance systems. Given the increasing adoption of ADAS and the growing interest in vehicle autonomy, the introduction of this distinguishing factor could lead to new advancements in scientific literature, triggering substantial improvements in road safety and traffic efficiency in emergency scenarios.

\section{Related Works}

In this section, we review several research lines in autonomous driving detection and emergency vehicle identification, including studies on fisheye camera datasets, audio-based detection methods, and traffic light recognition. While many excellent approaches exist in these areas, our review suggests that few studies have specifically addressed the problem using the particular combination of techniques we propose. This observation motivates our work and indicates that our method may offer a valuable complement—especially in scenarios where emergency vehicles do not activate their sirens—while remaining open to future multimodal integration.

\subsection{Fisheye Cameras and Datasets in Autonomous Driving}
Fisheye cameras are fundamental in computer vision applications for autonomous vehicles because they provide a wide panoramic view from a single sensor. However, this technology introduces strong radial distortions that affect the geometric representation of objects. Works such as \cite{Wu2023ADD:Driving} have developed datasets specifically designed for these conditions by incorporating calibration and distortion-correction techniques (e.g., models based on radial coefficients) that enable the generation of precise annotations for training detection models. In contrast, \cite{Li2020FisheyeDet:Images} addresses the challenge of adapting bounding box representations to the nonlinear geometry of fisheye images by exploring alternative representations (oriented boxes, ellipses, or polygons in polar coordinates) that maximize the \emph{Intersection of Union (IoU)} with the ground truth.

An important reference in this domain is the WoodScape dataset \cite{Yogamani2019WoodScape:Driving}. WoodScape is a comprehensive fisheye automotive dataset providing 360° coverage with four cameras and multi-task annotations. Its detailed semantic segmentation, depth estimation, and 3D object detection labels highlight the challenges—and the potential—of developing models that can operate directly on raw fisheye data without naive undistortion. In our work, we similarly leverage fisheye cameras and build on a detailed calibration process and an adapted data augmentation strategy using the Active Blue Light Dataset (ABLDataset), which has been developed as part of this study, to achieve highly accurate detection of blue emergency lights, thus overcoming the limitations imposed by fisheye distortions.

\subsection{Audio-Based Methods and Audio–Vision Fusion for Emergency Vehicle Detection}
Recent studies have investigated audio-based detection approaches for emergency vehicles. For example, Tran and Tsai \cite{Tran2020Acoustic-BasedNetworks} present a CNN-based system that leverages both raw waveform processing and frequency-domain features (such as MFCCs and log-mel spectrograms) to reliably classify siren signals even under adverse conditions. In a complementary work, Tran and Tsai \cite{Tran2021Audio-VisionDetection} propose an integrated audio–vision system that fuses information from both modalities, enhancing detection robustness when one modality is impaired.

It is important to recognize that there are situations where emergency vehicles may not activate their sirens, or where the siren is audible while the vehicle itself is not directly visible. In such scenarios, a visual detection system can serve as a critical complementary signal. Our approach—focused exclusively on the high-precision detection of blue emergency lights using a 360° multi-camera configuration and Color Attention RT-DETR (CA RT-DETR)—is fully capable of operating independently. Moreover, it is fully compatible with multimodal systems, where audio cues may complement and further enhance overall detection performance. 

\subsection{Detection of Lights in Road Infrastructure}
Numerous studies have focused on detecting luminous components in road infrastructure, particularly for traffic light recognition. For instance, \cite{Yao2023TL-Detector:Vehicles} presents a lightweight model for real-time traffic light detection that utilizes an enhanced backbone and attention modules to focus on localization features. Likewise, \cite{Lin2024TrafficIdentification} proposes a two-stage framework that detects individual signal bulbs, allowing recognition of not only standard circular lights but also more complex configurations such as arrow signals.

Unlike these works—which target traffic signals—our study is directed toward the detection of emergency lights, which exhibit unique chromatic and intensity properties that challenge conventional detectors. To address this, our approach builds on the RT-DETR architecture by integrating several key enhancements. In particular, we employ a Color Attention Module that leverages the input to emphasize regions corresponding to the distinctive blue hues—using chromatic masks in the HSV color space—and incorporates a residual connection to preserve essential information from the original image. Additionally, our method benefits from tailored calibration procedures and data augmentation strategies designed to effectively mitigate fisheye distortion. 

\subsection{Detection of Emergency Vehicles Using Vision and Neural Networks}
Training models to detect complete emergency vehicles is particularly challenging due to the substantial variability in appearance, configuration, and distribution of distinctive elements (such as lights, shapes, and emblems) among ambulances, fire trucks, and police vehicles. Various studies have tackled this problem using deep neural network architectures—such as YOLO or Faster R-CNN—that process the entire vehicle image. However, as noted in \cite{Shatnawi2024AnGAN}, models that rely on whole-vehicle detection often face generalization difficulties because of the intrinsic diversity of their visual features. For instance, the work by Shatnawi and Bani Younes demonstrates that augmenting the dataset with GANs can boost accuracy (exceeding 90\%) but also highlights the complexity of detecting vehicles in their entirety.

In contrast, our approach focuses on detecting the active blue lights as the primary indicator of an emergency vehicle, thereby optimizing the problem by concentrating on a critical visual component. Moreover, recent work on multi-vehicle tracking and counting \cite{GladiensyahBihanda2024Multi-VehicleByteTrack} shows that transformer-based detection architectures—such as RT-DETR—can robustly detect vehicles under challenging conditions (including occlusion and motion blur). This supports our decision to use RT-DETR backbone and a chromatic attention module in our system. By isolating the detection task to the specific feature of active blue lights, our method avoids the complexity of modeling the entire vehicle and achieves higher precision and recall in real-world conditions.

\section{Methodology} \label{sec:metho}
The primary objective of this study is to detect emergency vehicles by identifying activated blue lights. Specifically, we consider that detecting at least one light belonging to the same emergency vehicle is sufficient to determine that the vehicle has been identified. This hypothesis is based on the widespread use of blue lights as a priority signal in vehicles belonging to healthcare services, fire departments, police forces, and other security agencies.

\begin{figure}[!ht]
    \centering
    \includegraphics[width=\columnwidth]{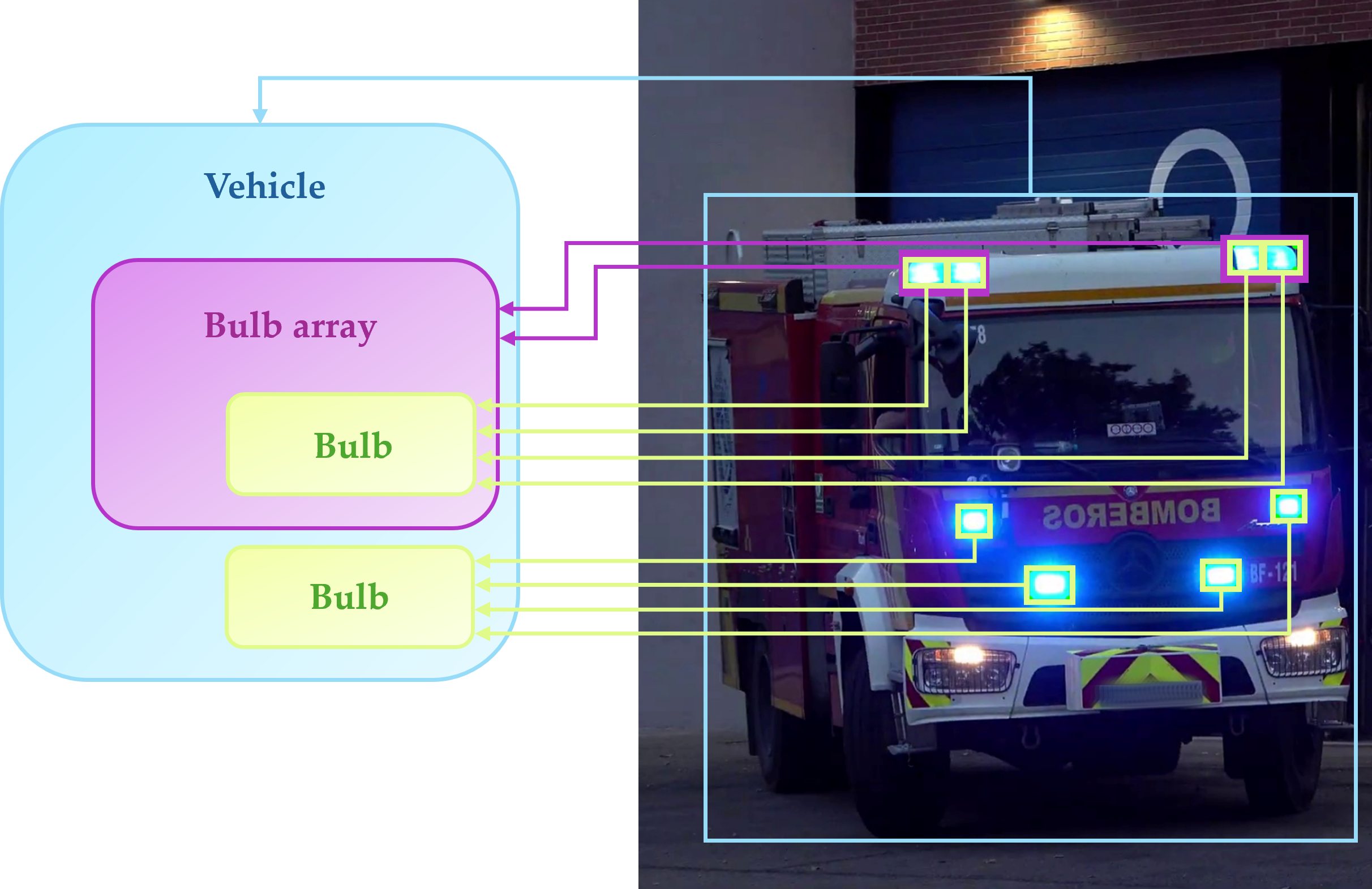}
    \caption{Levels of blue light grouping in emergency vehicles.}
    \label{fig:fig3}
\end{figure}

In terms of luminous structure, we distinguish several levels of grouping (Fig.~\ref{fig:fig3}):
\begin{itemize}
    \item \textbf{Bulb}: Each individual blue light source.
    \item \textbf{Bulb Array}: A set of contiguous or closely positioned lights functioning as a single unit, often housed within the same lighting enclosure.
    \item \textbf{Vehicle}: The total set of blue lights installed. A vehicle is considered detected if at least one of its bulbs is correctly identified within the image frame.
\end{itemize}

This detection strategy, based on the presence of active lights, simplifies the problem and ensures a high recall, as any illuminated blue light is, by definition, an indicator of the presence of an emergency vehicle. The following subsections describe how we constructed a dedicated dataset, the data augmentation techniques employed, the evaluation of different detection models, and a context-aware approach to analyzing metrics.

\subsection{ABLDataset}
\label{sec:abldataset}
As previously mentioned, no specialized dataset for the detection of active blue emergency lights existed. For this reason, we created the \textit{Active Blue Light Dataset (ABLDataset)}, consisting of annotations on images extracted from YouTube videos. This strategy follows approaches used, partially or entirely, by other datasets such as YouTube-BoundingBoxes \cite{Real2017YouTube-BoundingBoxes:Video}, BDD100K \cite{Yu2020BDD100K:Learning}, and TAO \cite{Dave2020TAO:Object}. By leveraging a diverse set of online video sources, we obtained a dataset rich in scenarios, vehicles, lighting structures, and weather conditions.

We owe special thanks to individuals who record and publish videos of emergency vehicles, also known as emergency vehicle spotters \cite{UrgencesGeneve2024UrgencesGeneveYouTube,JuanEmergencias2024juanemergenciasYouTube}. Their work has produced a wealth of high-quality data for training neural networks.

Our approach focuses on creating a single class, named \emph{Active Blue Light}, referring exclusively to blue lights that are turned on or flashing. This single-class strategy:
\begin{itemize}
    \item Simplifies the training process, allowing the model to be optimized for one specific target;
    \item Achieves faster processing speeds and higher accuracy, as the model focuses on a consistent chromatic feature across many emergency vehicles;
    \item Reduces computational complexity and inference time, both of which are critical for real-time detection in traffic scenarios.
\end{itemize}

\subsubsection{Dataset Characteristics}
\begin{figure}[!ht]
    \centering
    \includegraphics[width=\columnwidth]{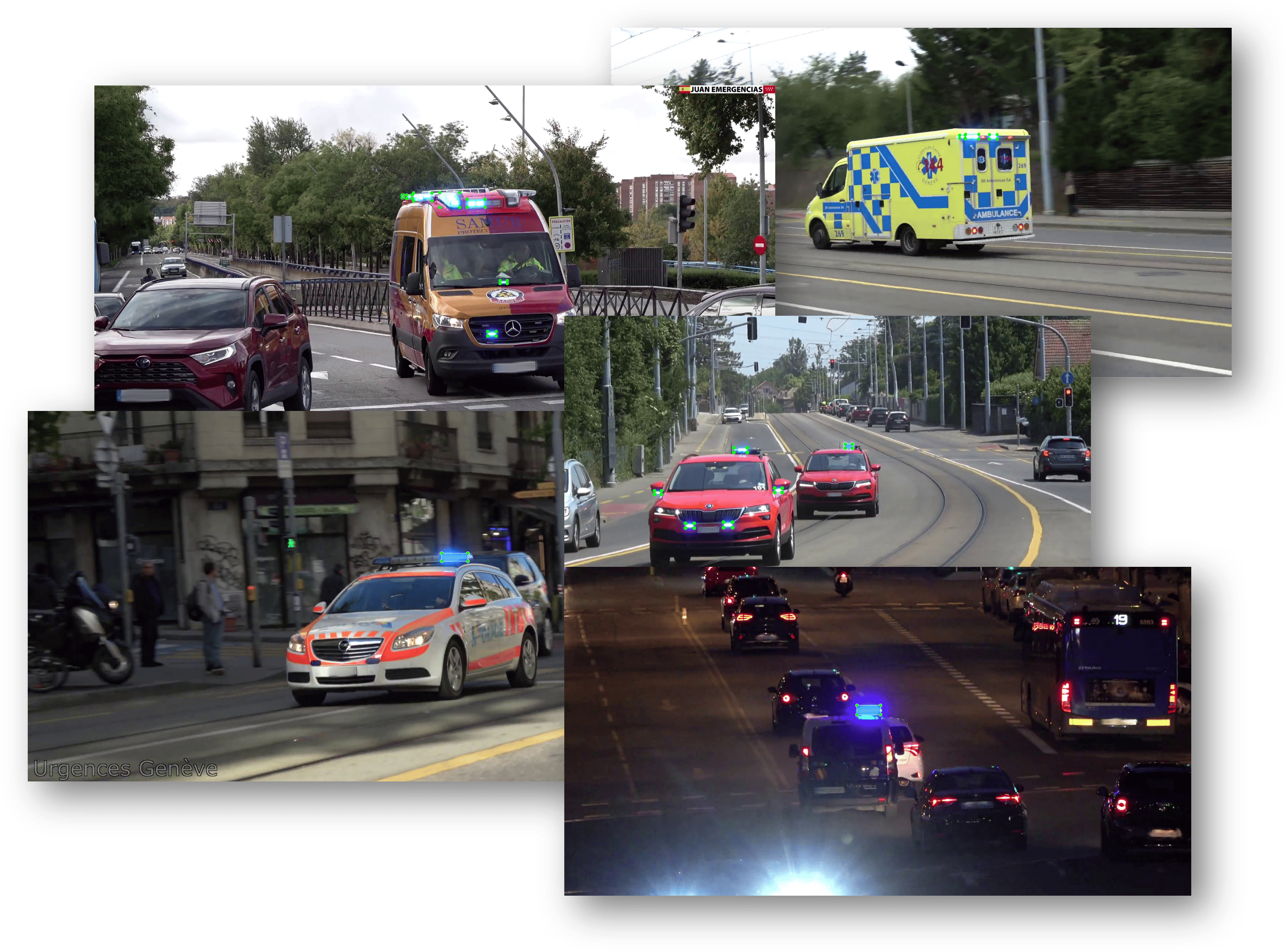}
    \caption{Examples of labeled images from the ABLDataset.}
    \label{fig:fig4}
\end{figure}

The dataset comprises about 3{,}000 images of various traffic situations featuring emergency vehicles with active blue lights (Fig.~\ref{fig:fig4}). The captured frames include vehicles from diverse agencies in Europe (Spain, the U.K., Portugal, Italy, Hungary, and Germany).

For annotation, we adopted a series of standardization criteria:
\begin{enumerate}
    \item The bounding box must encompass only the bright, active portion of the blue light, stopping before the background starts to become visible.
    \item For devices with multiple bulbs, two scenarios arise:
          \begin{itemize}
              \item Adjacent bulbs sharing a bright, unified light are annotated as a single bounding box.
              \item If a bluish area between bulbs is faint and background color appears, they are annotated in separate boxes.
          \end{itemize}
    \item The bounding box should closely match the light source shape as seen by the camera.
\end{enumerate}
These guidelines proved efficient for annotation and produced strong results during training.

\subsubsection{Dataset Distribution and Management}
The 3{,}049 images contain 10{,}437 annotations in total, and are divided into:
\begin{itemize}
    \item \texttt{train} (69.3\%): 2112 images, 7208 annotations
    \item \texttt{val} (15.8\%): 482 images, 1586 annotations
    \item \texttt{test} (14.9\%): 455 images, 1643 annotations
\end{itemize}

\begin{figure}[!ht]
    \centering
    \includegraphics[width=\columnwidth]{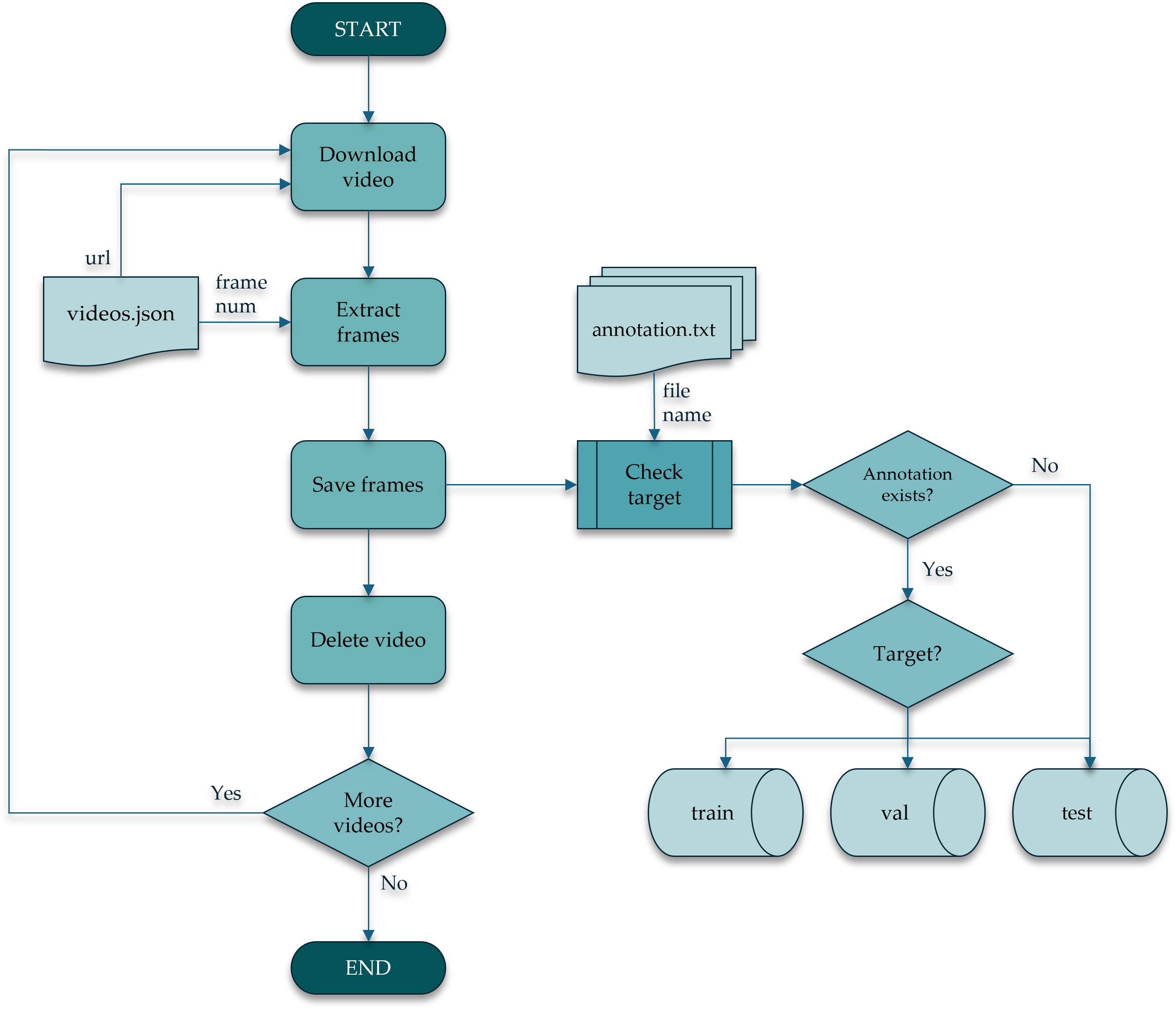}
    \caption{Flowchart of the dataset creation process.}
    \label{fig:fig5}
\end{figure}

A specialized software tool was developed to automate video downloads, frame extraction, and distribution of images among subsets (Fig.~\ref{fig:fig5}). All annotations are stored in the YOLO format \cite{UrgencesGeneve2024UrgencesGeneveYouTube}, with optional converters for CreateML \cite{Inc.2018CreateDocumentation} and PascalVOC \cite{Everingham2010TheChallenge}. The annotations are made publicly available via a dedicated GitHub repository.

\subsubsection{Data Augmentation} \label{subsec:augmentation}
Considering that the purpose of this dataset is to serve as the training source for a detection system whose images will be captured by built-in fisheye cameras in automobiles, the images were manipulated to adapt to the conditions under which inference will ultimately occur. This process involved calibrating actual fisheye cameras to obtain intrinsic parameters and distortion coefficients.

The intrinsic calibration of the fisheye camera lens begins by acquiring a set of images in which a chessboard pattern is placed at various positions and orientations. From each image, corners are detected using feature recognition algorithms, associating the 2D positions with the 3D coordinates of the calibration pattern in the real world. Subsequently, an iterative adjustment process refines both the intrinsic matrix, which includes the focal lengths $(f_x, f_y)$ and the principal point $(c_x, c_y)$, and the distortion coefficients. Unlike the simpler radial and tangential distortion models typical of pinhole lenses \cite{BROWNDC1971Close-Calibration}, a specialized fisheye distortion model is used \cite{Kannala2006ALenses}, employing four radial coefficients $(k_1, k_2, k_3, k_4)$ to describe the pronounced lens deformations.

\begin{figure}[!ht]
    \centering
    \includegraphics[width=\columnwidth]{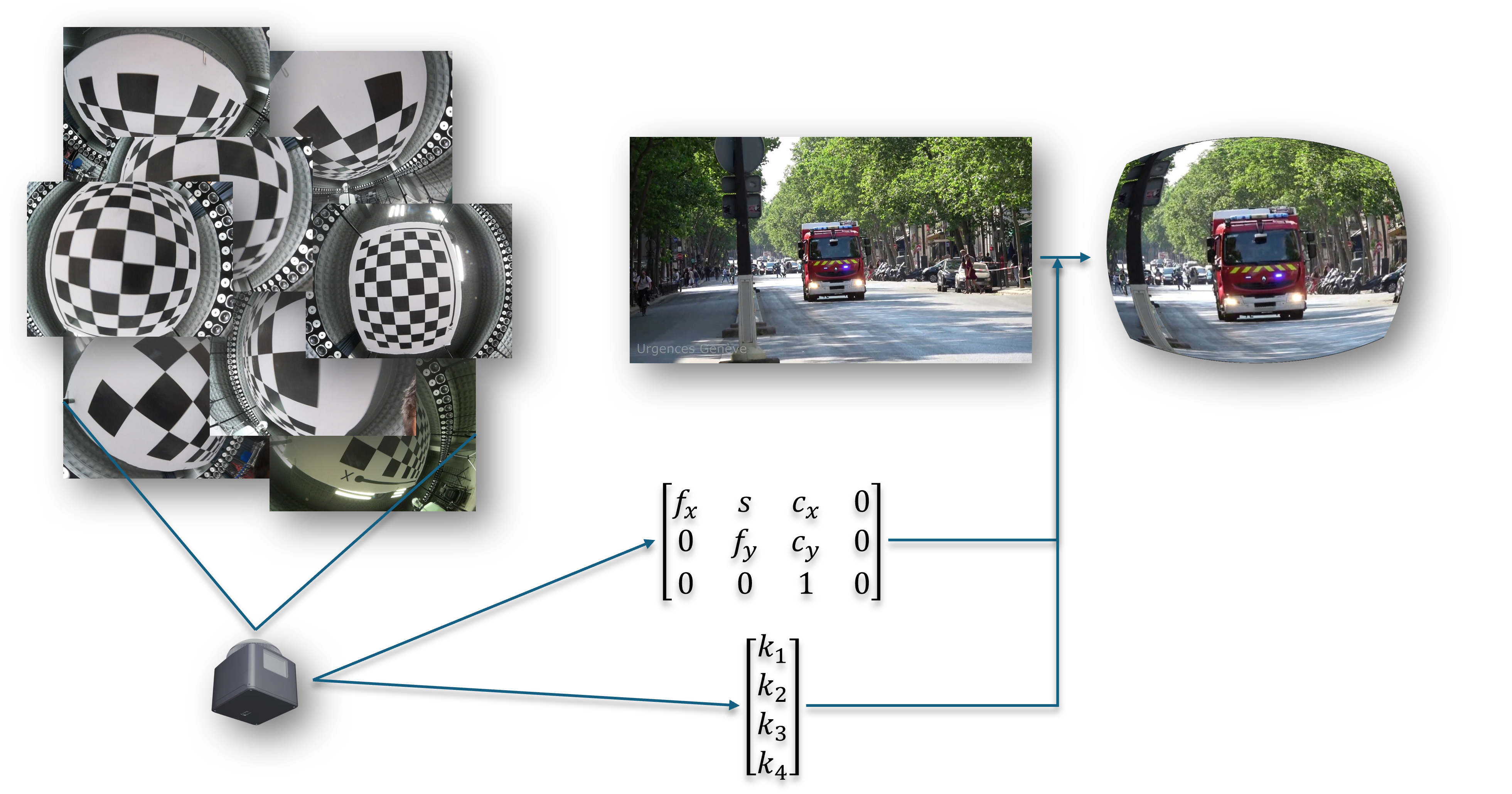}
    \caption{Explanatory diagram of the lens distortion procedure for data augmentation.}
    \label{fig:fig6}
\end{figure}

Once these parameters are known, the dataset images undergo a process of simulated lens distortion with a random variation of less than 10\% on the distortion coefficients, thereby generalizing the result to cameras with slightly different coefficients. Figure~\ref{fig:fig6} summarizes the workflow. Applying lens distortion to the dataset also requires transforming the coordinates of the bounding boxes. Following the approach of Li \emph{et al.} \cite{Li2020FisheyeDet:Images}, the bounding box edges are kept parallel, applying the distortion function to the unnormalized bounding box coordinates so that they correctly shift in the distorted image.

Additionally, several minor modifications enhance the generalizability of training, simulating variations in perspective, scale, illumination, noise, and rotation \cite{Shorten2019ALearning}:
\begin{itemize}
    \item Random rotations within a range of $\pm 15^\circ$,
    \item Perspective changes with a factor of 0.0001,
    \item Uniform scaling between 0.8 and 1.2,
    \item Brightness and contrast adjustments,
    \item Gaussian noise to simulate adverse conditions.
\end{itemize}
These transformations help ensure that the system remains robust to moderate variations in viewpoint and illumination once deployed in real-world vehicular environments.

\subsubsection{Standard Dataset Evaluation}
\label{subsec:std_dataset_eval}
To analyze the quality of the dataset and identify the model with the best performance for this data, a comprehensive comparison of state-of-the-art object detection models was conducted. These models are widely utilized in various industrial applications and have significantly enhanced driver assistance systems in vehicles \cite{Wang2023VV-YOLO:YOLOv4,Yatbaz2024Run-TimeRepresentations,Dazlee2022ObjectYOLO}. 

All training and validation processes were performed using an NVIDIA RTX 4090 GPU, ensuring efficient processing of the dataset and associated deep learning models.

The models employed to evaluate the dataset's performance were as follows:
\begin{itemize}
    \item \textbf{RetinaNet}: Based on a Feature Pyramid Network (FPN) built on a ResNet backbone, it incorporates a focal loss to penalize majority classes and balance training on imbalanced datasets \cite{Lin2017FocalDetection}.
    \item \textbf{Faster R-CNN}: Combines a Region Proposal Network (RPN) to generate high-quality candidates with a detection network that classifies and adjusts bounding boxes, using a shared convolutional backbone \cite{Ren2015FasterNetworks}.
    \item \textbf{YOLOv5}: A deep convolutional network that uses a detection grid to make direct predictions of classes and bounding boxes, optimized for fast inference times \cite{Hussain2024YOLOv5Vision}.
    \item \textbf{YOLOv8}: Integrates dynamic anchor modules and multiscale feature fusion, enhancing computational efficiency and spatial representation for real-time detection tasks \cite{Hussain2024YOLOv5Vision}.
    \item \textbf{YOLOv10}: Optimizes its convolutional backbone and multiscale feature fusion to improve the detection of small objects without sacrificing inference speed. It also incorporates regularization adjustments and resolution scaling to facilitate convergence and training on imbalanced datasets \cite{Hussain2024YOLOv5Vision,Wang2024YOLOv10:Detection}.
    \item \textbf{RT-DETR}: Employs a transformer architecture with an encoder-decoder to process features extracted from images, capturing long-range spatial relationships through attention mechanisms \cite{Zhao2024DETRsDetection}.
\end{itemize}

The evaluation results on the test set are summarized in Table~\ref{tab:table1}.

\begin{table}[!ht]
\centering
\caption{Preliminary comparison of standard metrics for different object detection models (no modifications). These results guide the selection of the baseline model.}
\label{tab:table1}
\resizebox{\columnwidth}{!}{%
\begin{tabular}{|l|c|c|c|c|c|}
\hline
\rowcolor[HTML]{9BB1AF} 
\cellcolor[HTML]{C8DAD5}\textbf{Model} & \cellcolor[HTML]{9BB1AF}\textbf{P(\%)} & \cellcolor[HTML]{9BB1AF}\textbf{R(\%)} & \cellcolor[HTML]{9BB1AF}\textbf{mAP$_{0.5}$} & \cellcolor[HTML]{9BB1AF}\textbf{mAP$_{0.5:0.95}$} & \cellcolor[HTML]{9BB1AF}\textbf{$t_i$ (ms)} \\
\hline
Retinanet   & 61.3 & 69.3 & 57.8 & 31.0 & 5.3 \\
\hline
Faster R-CNN & 77.3 & 75.2 & 80.1 & 49.2 & 5.7 \\
\hline
Yolov5 -- M  & 81.9 & 75.6 & 82.5 & 51.8 & 2.0 \\
\hline
Yolov8 -- M  & 85.8 & 72.7 & 82.4 & 50.9 & 2.2 \\
\hline
Yolov10 -- M & 83.2 & 70.8 & 80.6 & 50.5 & 2.4 \\
\hline
RT-DETR      & 85.0 & 82.8 & 89.1 & 53.2 & 3.7 \\
\hline
\end{tabular}%
}
\end{table}

The results presented in the table reveal several interesting trends among the analyzed models. For instance, Faster R-CNN, despite being a two-stage method (which typically results in longer inference times, $t_i$), delivers strong performance in terms of precision and mean Average Precision (mAP). However, it falls short of achieving the recall levels reached by RT-DETR. On the other hand, RetinaNet, which leverages the well-known focal loss to address data imbalances, achieves acceptable results but underperforms in the main metrics, translating into a reduced ability to detect the majority of blue lights compared to other architectures.

The YOLO variants (v5, v8, v10) stand out for their computational efficiency and fast inference times, two factors of significant importance in embedded environments or real-time applications. However, it is observed that in this specific case, their recall tends to be somewhat lower than that of RT-DETR, implying a higher risk of missing some emergency lights. Additionally, the more recent versions, such as YOLOv8 and YOLOv10, incorporate improvements in multiscale feature fusion and regularization adjustments, achieving more balanced metrics compared to YOLOv5. However, they still fall short of surpassing the trade-off between precision and recall provided by RT-DETR.

In summary, while several models are competitive—some prioritizing speed and others focusing on detection accuracy—RT-DETR emerges as the best overall option. It combines higher recall with robust precision, which is critical in detecting blue emergency lights where a high false-negative rate cannot be tolerated.

\subsubsection{Contextualized Evaluation of the Dataset}
\label{subsec:context_eval}
Given the specific nature of detecting vehicles with active emergency lights, we developed customized metrics. As noted earlier (Section~\ref{sec:metho}), different levels of grouping were defined to contextualize the problem, considering the presence of a blue light as indicative of an emergency vehicle.

\textbf{Bulb Array Precision} ($P_{BA}$) measures the fraction of correct predictions over the total number of predictions made. A prediction is considered correct (true positive, TP) if it overlaps with a ground truth bounding box, and each ground truth bounding box can contribute to only one TP:
\begin{equation}
    P_{BA} = 
    \frac{\sum_{i=1}^{M} \min\!\Bigl(1, \sum_{j=1}^{N} I(\mathrm{IoU}(\mathrm{pred}_j,\mathrm{gt}_i) > \tau)\Bigr)}{M},
\end{equation}
where:
\begin{itemize}
    \item $N$ is the number of ground truth elements,
    \item $M$ is the number of predicted elements,
    \item $I(\cdot)$ is an indicator function that equals 1 if the condition is true and 0 otherwise,
    \item $\mathrm{IoU}(\cdot)$ calculates the Intersection over Union between two bounding boxes,
    \item $\tau$ is a predefined IoU threshold (0.5) that determines whether a predicted bounding box sufficiently overlaps with a ground truth box.
\end{itemize}

\begin{figure}[!ht]
    \centering
    \includegraphics[width=0.9\columnwidth]{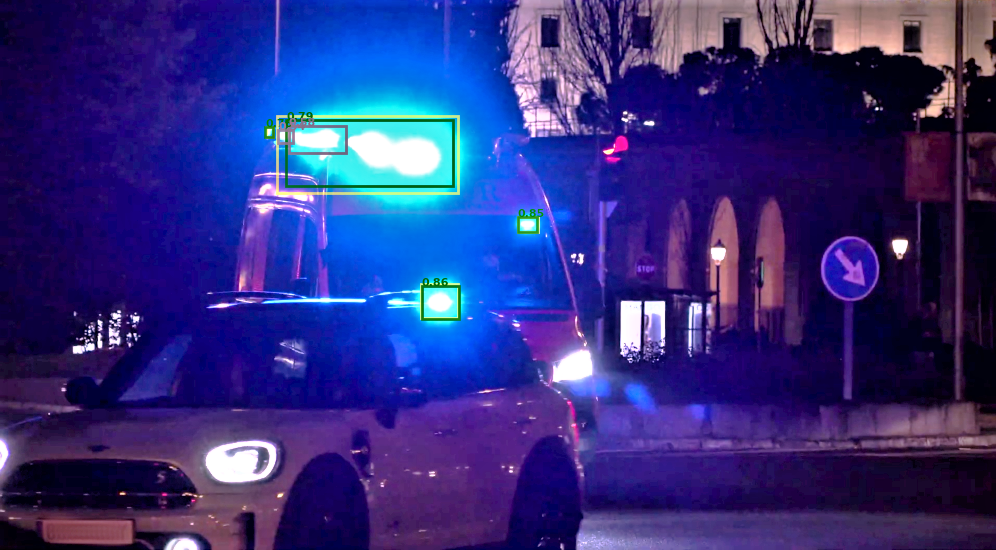}
    \caption{Example of overlapping detections on the same ground truth.}
    \label{fig:fig7}
\end{figure}

This metric better reflects how effectively inferences handle the lighting structures in emergency vehicles, which may contain multiple bulbs within the same enclosure. Detecting any portion of the array is considered valid, since an overlap with at least one light in the structure suffices.

Similarly, \textbf{Bulb Array Recall} ($R_{BA}$) measures the fraction of correct predictions over the total number of ground truth bounding boxes. A prediction is correct (true positive) if it overlaps with a ground truth bounding box:
\begin{equation}
    R_{BA} =
    \frac{\sum_{i=1}^{N} \min\!\Bigl(1, \sum_{j=1}^{M} I(\mathrm{IoU}(\mathrm{pred}_j,\mathrm{gt}_i) > \tau)\Bigr)}{N}.
\end{equation}
The same definitions for $N$, $M$, $I(\cdot)$, $\mathrm{IoU}(\cdot)$, and $\tau$ apply here as well.

Because a lighting structure can comprise multiple bulbs, an emergency vehicle can typically have several such structures. Therefore, additional metrics are introduced to assess vehicle-level detection performance. To facilitate this evaluation, in addition to the \texttt{train}, \texttt{val}, and \texttt{test} directories, a fourth directory named \texttt{test\_id} was created. For each file in the test set, this directory contains a complementary file that records—row by row—whether a given annotation belongs to the same vehicle or different ones within each image.

\begin{figure}[!ht]
    \centering
    \includegraphics[width=\columnwidth]{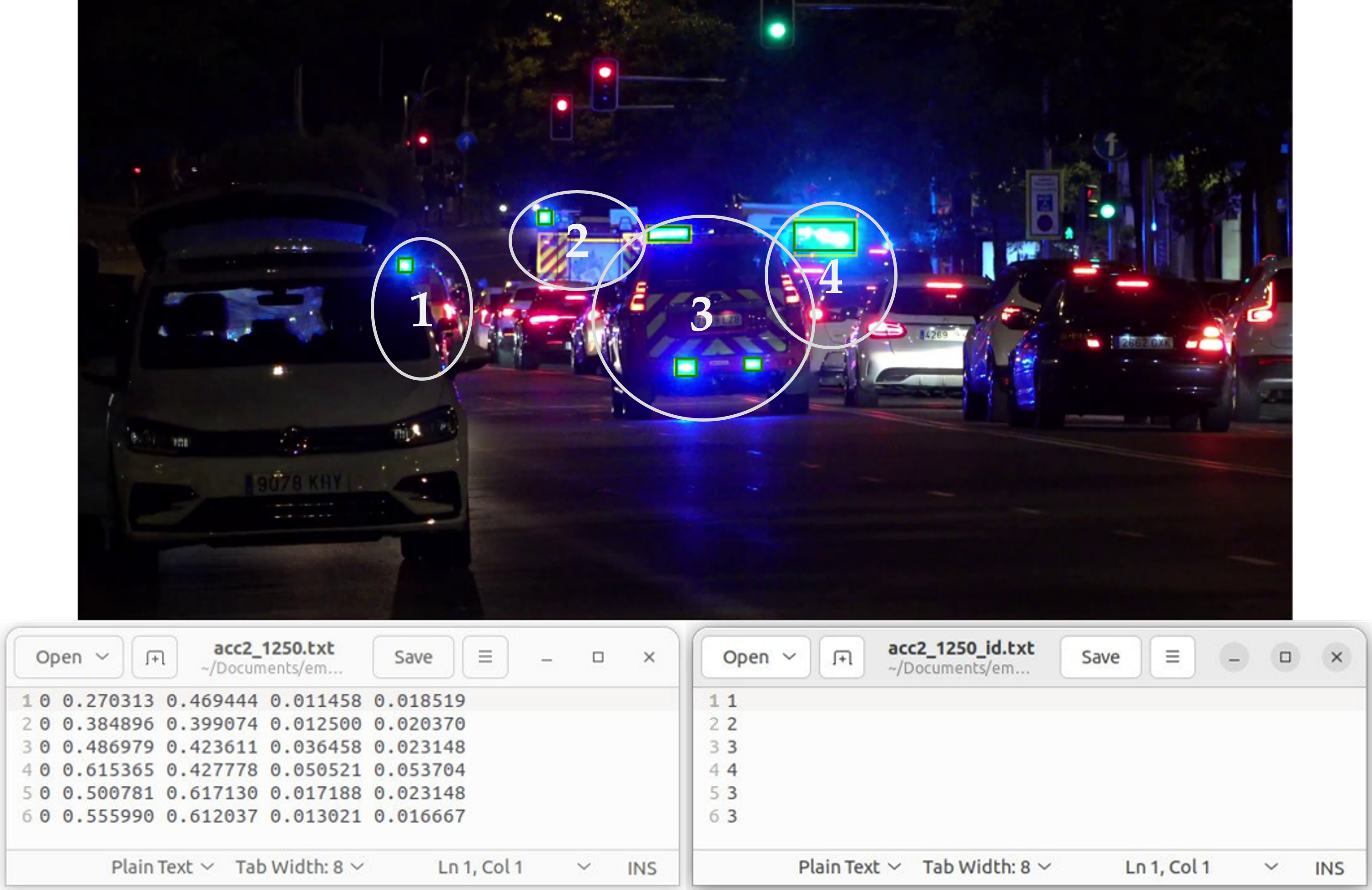}
    \caption{Example of vehicle-based annotation. The image shows a scene where four emergency vehicles are present, each labeled as a distinct vehicle in the dataset with its corresponding bounding box annotations.}
    \label{fig:fig8}
\end{figure}

\textbf{Vehicle Precision} ($P_V$) measures the fraction of correctly detected vehicles over the total number of \emph{predicted} vehicles. A vehicle is considered correctly detected (true positive vehicle, TP$_V$) if at least one of its annotations overlaps with a prediction:


\begin{equation}
\label{eq:pv}
\resizebox{0.95\columnwidth}{!}{$
P_V = \frac{
  \sum_{k=1}^{K} 
    \min\!\bigl(
      1,\,
      \sum_{l=1}^{L} 
      \sum_{u=1}^{n_l} 
      \sum_{v=1}^{m_k} 
      I\bigl(\mathrm{IoU}(\mathrm{pred}_{(l,u)}, \mathrm{gt}_{(k,v)})>\tau\bigr)
    \bigr)
}{L}.
$}
\end{equation}

where:
\begin{itemize}
    \item $K$ is the total number of ground-truth vehicles,
    \item Each vehicle $k$ contains $m_k$ annotated lights: $\mathrm{gt}_{(k,1)}, \dots, \mathrm{gt}_{(k,m_k)}$,
    \item $L$ is the number of \emph{predicted} vehicles,
    \item Each predicted vehicle $l$ groups $n_l$ bounding boxes of lights: $\mathrm{pred}_{(l,1)}, \dots, \mathrm{pred}_{(l,n_l)}$.
\end{itemize}
Here, $\mathrm{IoU}(\mathrm{pred}_{(l,u)},\mathrm{gt}_{(k,v)})$ is again the intersection over union, and $\tau$ the threshold.

Finally, \textbf{Vehicle Recall} ($R_V$) measures the fraction of correctly detected vehicles over the total number of \emph{actual} vehicles in the ground truth. A vehicle is considered correctly detected if at least one of its bulbs overlaps with a predicted box:
\begin{equation}
\label{eq:rv}
\resizebox{0.95\columnwidth}{!}{$
R_V =
\frac{
\sum_{k=1}^{K} 
  \min\!\Bigl(
    1,\,
    \sum_{l=1}^{L}
    \sum_{u=1}^{n_l}
    \sum_{v=1}^{m_k}
      I\bigl(\mathrm{IoU}(\mathrm{pred}_{(l,u)},\,\mathrm{gt}_{(k,v)}) > \tau\bigr)
  \Bigr)
}{K}.
$}
\end{equation}
These additional metrics provide a more holistic understanding of how effectively the model detects an entire vehicle, rather than focusing solely on individual bulbs or arrays.

\subsection{Color Attention RT-DETR}
\label{subsec:colorAttn}
The adaptation of models by incorporating specialized blocks that align with the nature of the problem is a widely used technique \cite{Yu2024MCG-RTDETR:Imagery,Zhang2024RS-DETR:RT-DETR}. Emergency light detection heavily relies on identifying shades of blue and white with bluish tints. To maximize the performance of RT-DETR in this context, a Color Attention Module has been integrated, enhancing chromatic regions of interest before the primary feature extraction and detection stage (Figure \ref{fig:fig9}). This strategy is based on the following scientific justification:

\begin{figure*}[!ht]
    \centering
    \includegraphics[width=\textwidth]{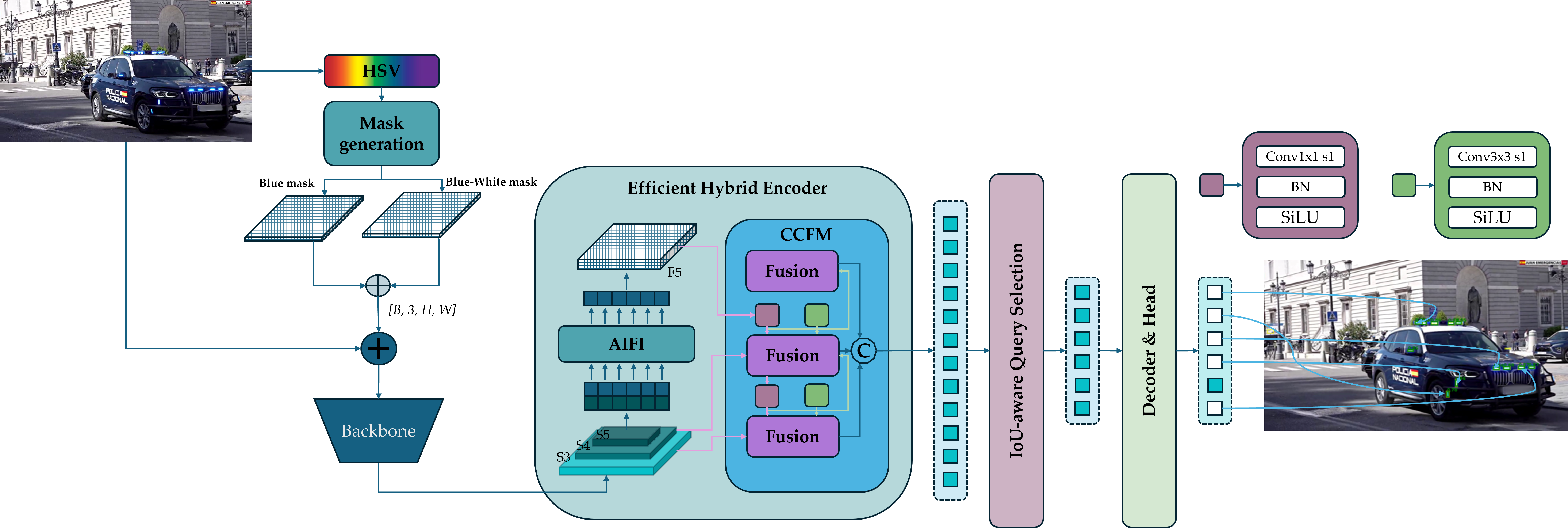}
    \caption{Color Attention - RT-DETR Diagram.}
    \label{fig:fig9}
\end{figure*}

\paragraph*{Justification Based on Color Theory and Advantages of the HSV Color Space}
From a physiological and perceptual standpoint, the color blue exhibits high contrast against urban backgrounds and environments dominated by grays and blacks. However, many emergency lights include intense white components with subtle blue tints, which can confuse generic detectors. To handle this variability, the HSV (Hue-Saturation-Value) color space is employed, facilitating the isolation of specific color regions (hue) and the discrimination of intensity (value). Appropriate ranges are defined to capture both pure blue pixels and high-luminance pixels that exhibit a shift towards bluish tones.

\paragraph*{Chromatic Attention Block and Residual Connection}
After converting the image to the HSV color space, two independent masks are applied:
\begin{itemize}
    \item One targeting pure blue regions
    \item One capturing whites with a blue component
\end{itemize}
To minimize the alteration of potentially relevant visual features in other regions, the block incorporates a residual sum connection:
\begin{equation}
    \text{output} = x + \bigl(x \times \text{mask}\bigr),
\end{equation}
where $x$ represents the normalized image and \texttt{mask} is the combination of the two chromatic ranges. This mechanism enhances pixels associated with emergency lights while preserving the general object detection capabilities of RT-DETR.

\paragraph*{Integration into the RT-DETR Architecture}
Figure~\ref{fig:fig9} illustrates the insertion of the Color Attention Module before processing through the transformer in RT-DETR. With this adjustment, the model learns to focus its attention on areas where the blue component---whether pure or whitish---is predominant. This leads to better discrimination of emergency light signals in complex scenarios or under varying lighting conditions.


This modification increases the model’s robustness to variations in light intensity and hue, enabling reliable detection of emergency vehicles even when the color deviates from the ideal blue. Consequently, detection at medium and long distances is notably improved, yielding consistently higher precision and recall.

\subsection{Multi-Camera System}
\label{subsec:multicamera}
As mentioned in the Introduction (Section~\ref{sec:intro}), a multi-camera system equipped with fisheye lenses has been developed to validate the model's functionality and demonstrate a product that could be integrated into current ADAS (Advanced Driver Assistance Systems). This system follows the camera arrangement used in the WoodScape Dataset \cite{Yogamani2019WoodScape:Driving}, contributing to its potential expansion by incorporating new annotations for blue emergency lights.

\begin{figure}[!ht]
    \centering
    \includegraphics[width=\columnwidth]{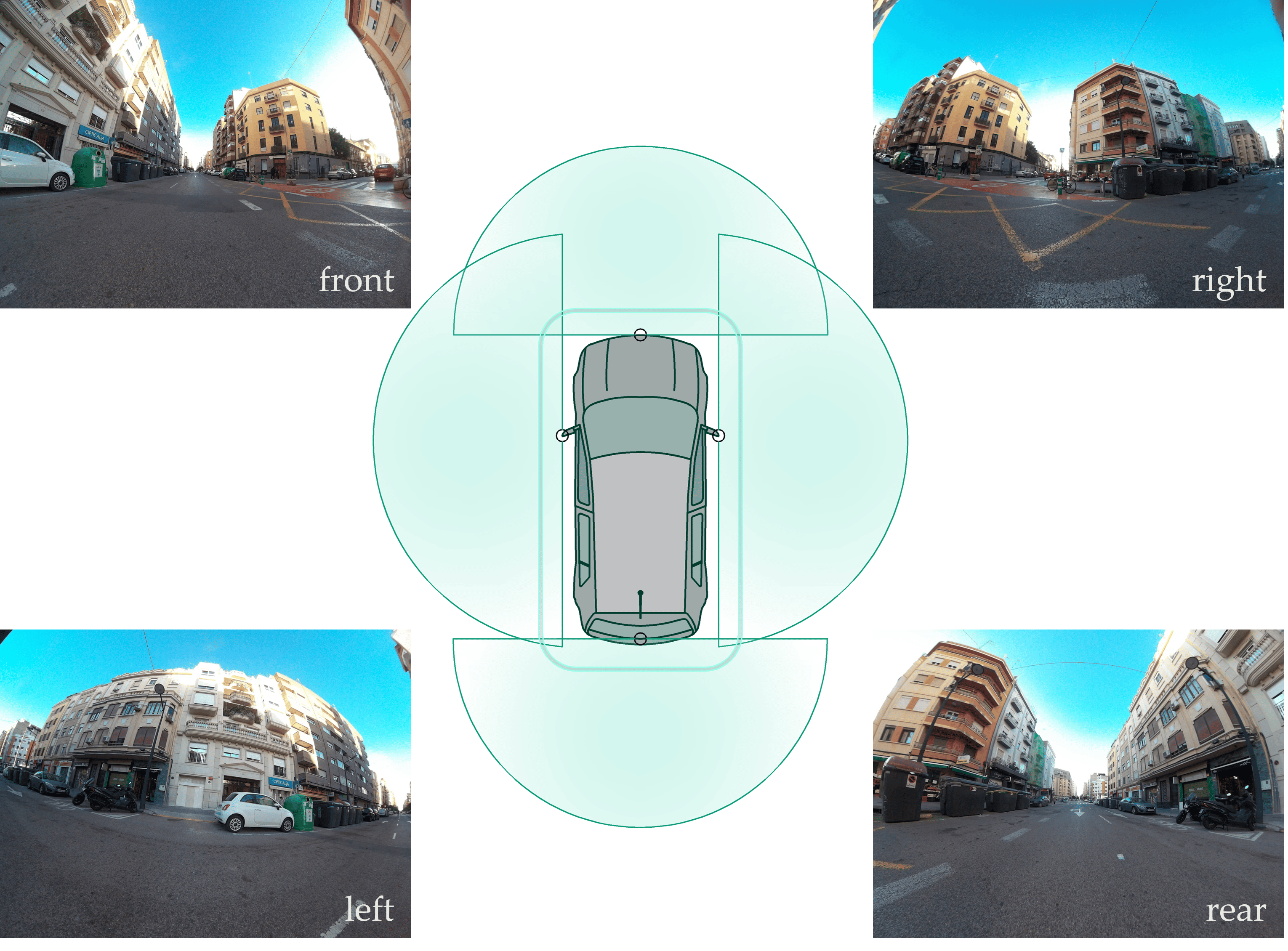}
    \caption{Multi-Camera Fisheye System Configuration for 360\degree{} Coverage.}
    \label{fig:fig10}
\end{figure}

The system consists of four fisheye cameras positioned at the front, side mirrors, and rear windshield of the vehicle, as illustrated in Fig.~\ref{fig:fig10}. This setup, along with a 180\degree{} horizontal field of view (FoV), ensures full 360\degree{} coverage of the vehicle’s surroundings. Furthermore, this configuration---with slight variations in the exact positioning of the cameras---is widely adopted in the automotive industry \cite{Horgan2015Vision-BasedAdvances}.


To determine the detection angle of emergency vehicles, a camera calibration process was carried out to obtain their intrinsic parameters as well as the relative position and orientation between them (extrinsic parameters) with respect to the central point of the vehicle. This calibration enables the creation of a transfer function that converts the pixel coordinates of detections in each camera into an azimuthal angle, providing valuable information for the driver.

\subsubsection{Calibration}
\label{subsubsec:calibration}
The calibration system consists of two main steps: first, obtaining the intrinsic parameters and distortion coefficients of the individual cameras using the procedure described in Section~\ref{subsec:augmentation}, and second, determining the extrinsic parameters to establish the relative position and orientation of the cameras.

The following describes the extrinsic calibration process used to determine the relative position and orientation of the different fisheye cameras on the vehicle, as well as the method to obtain the azimuthal detection angle from these geometric relationships. This methodology follows the principles of multi-camera system calibration based on overlapping fields of view, as detailed in \cite{Knorr2014ExtrinsicView}, and relies on the joint estimation of rotation and translation transformations that unify each camera into a common reference frame (the vehicle's center).

To transform a 3D point $X_i^{(\mathrm{cam})}$, expressed in the coordinate system of camera~$i$, into the vehicle coordinate system $X^{(\mathrm{veh})}$, the rotation matrix $R_i$ and translation vector $T_i$ of camera~$i$ relative to this reference frame are required:
\begin{equation}
    X^{(\mathrm{veh})} = R_i \, X_i^{(\mathrm{cam})} + T_i.    
\end{equation}

In a system with $N$ cameras, the parameters $\{R_i,\,T_i\}$ must be estimated for $i \in \{1, \dots, N\}$. To obtain these parameters with high reliability despite the distortion introduced by fisheye lenses, a chessboard pattern was used as a reference for each camera pair. This pattern facilitates the robust detection of corners and the creation of initial point correspondences.

To further improve precision, extrinsic calibration was performed in both directions: not only is camera~$i$ calibrated relative to camera~$j$, but camera~$j$ is also calibrated relative to camera~$i$. This bidirectional approach ensures consistent rotation and translation estimates, increasing the overall robustness of the process.

Due to the strong fisheye distortion, a rectification stage is applied, where each image is virtually projected after correcting for intrinsic distortion using $K_i$ and $D_i$. Once rectified, homographies are used to align overlapping areas and extract/describe features using SIFT~\cite{Lowe2004DistinctiveKeypoints}, enabling the identification of corresponding points between adjacent cameras.

The final estimation of the extrinsic parameters $\{R_i,\,T_i\}$ is achieved by minimizing error functions based on the reprojection of these corresponding points, using a Levenberg-Marquardt optimizer within a bundle adjustment framework. The intrinsic calibration (matrix $K_i$ and coefficients $D_i$) had already been performed individually, as described in Section~\ref{subsec:augmentation}, using the same chessboard pattern for corner detection.

\subsubsection{Azimuthal Angle Estimation From Calibration}
Once the parameters $\{R_i,\,T_i\}$ have been estimated, any detected pixel $p = (u,v)$ in the image of camera~$i$ can be converted into a 3D vector in the vehicle's coordinate system. This enables the calculation of the azimuthal angle in the horizontal plane $(X, Y)$, indicating the direction relative to the vehicle's front axis.

\begin{figure}[!ht]
    \centering
    \includegraphics[width=0.7\columnwidth]{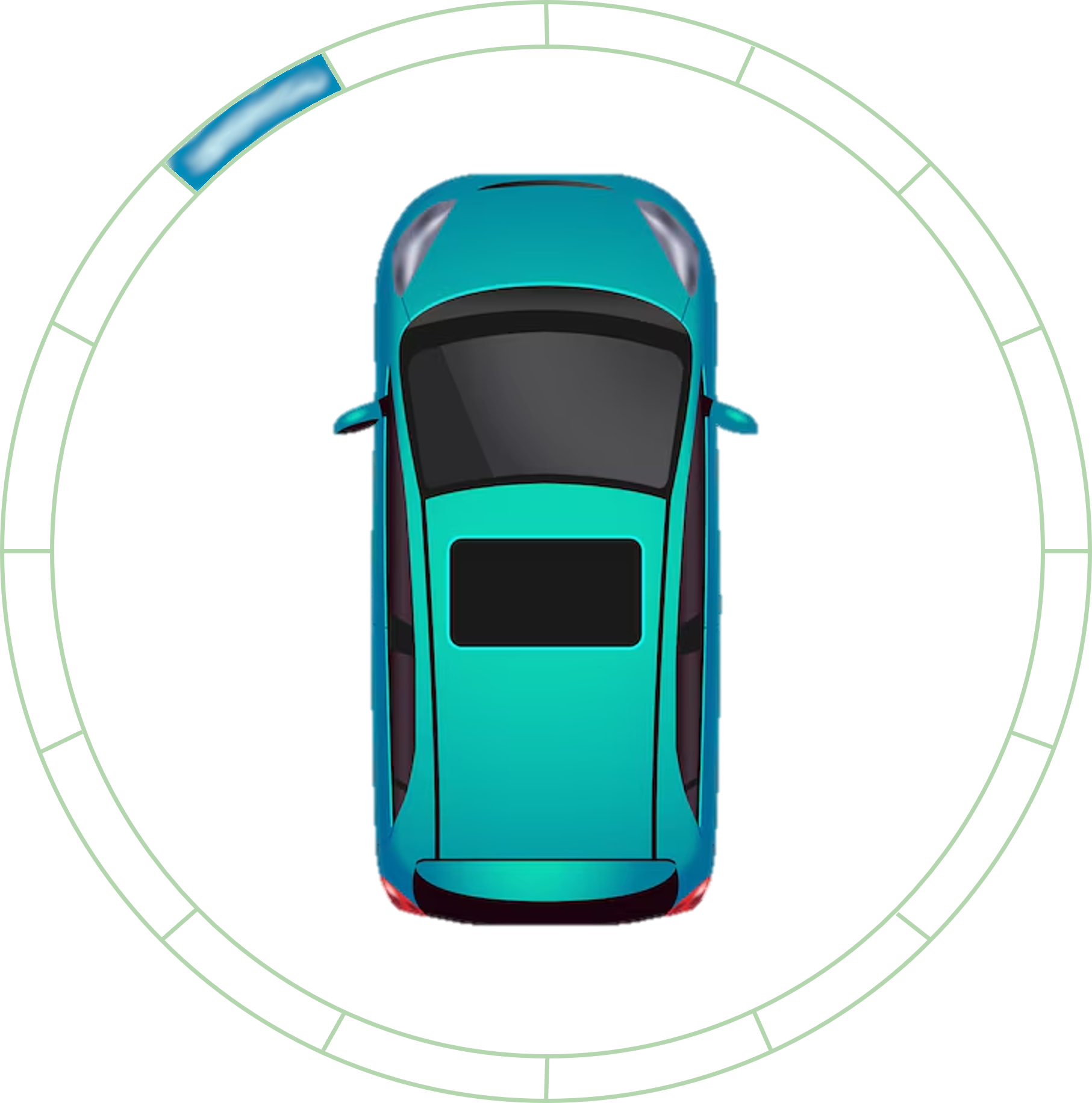}
    \caption{Example of a dashboard alert design for the driver, indicating the presence of an emergency vehicle.}
    \label{fig:fig11}
\end{figure}

First, a distortion correction and an inverse projection are performed. Using $K_i$ and $D_i$, the distortion of $p$ is corrected to find the direction ray $\tilde{x}_i$ in camera~$i$'s coordinates:
\begin{equation}
    \tilde{x}_i = f_{\mathrm{inv}}\bigl(K_i^{-1}(u,\,v,\,1)^\mathsf{T},\,D_i \bigr).
\end{equation}
Next, the transformation to vehicle coordinates is carried out: we multiply by $R_i$ and normalize,
\begin{equation}
    \tilde{x}^{(\mathrm{veh})} = R_i \,\tilde{x}_i.    
\end{equation}
Finally, the azimuthal angle $\varphi$ w.r.t.\ the $X$-axis of the vehicle is:
\begin{equation}
    \varphi = \arctan2(y,\;x)\,\times\,\frac{180}{\pi}.    
\end{equation}
This result directly indicates the orientation from which the emergency light is detected, allowing information to be fused with other sensors or to alert the driver about the emergency vehicle's location. Figure~\ref{fig:fig11} illustrates a possible design of the user interface for the driver.

\section{Results}
\subsection{Standard Results of the Proposed Network}
\label{subsec:stdResults}
The following presents the results obtained by the architecture incorporating the chromatic attention block (Color Attention RT-DETR), focused on the detection of blue emergency lights. Table~\ref{tab:table2} summarizes the key metrics:

\begin{table}[!ht]
\centering
\caption{Standard (non-contextualized) metrics for the customized model.}
\label{tab:table2}
\resizebox{\columnwidth}{!}{%
\begin{tabular}{|l|c|c|c|c|c|}
\hline
\rowcolor[HTML]{9BB1AF} 
\cellcolor[HTML]{C8DAD5}\textbf{Model} & \cellcolor[HTML]{9BB1AF}\textbf{P(\%)} & \cellcolor[HTML]{9BB1AF}\textbf{R(\%)} & \cellcolor[HTML]{9BB1AF}\textbf{mAP$_{0.5}$} & \cellcolor[HTML]{9BB1AF}\textbf{mAP$_{0.5:0.95}$} & \cellcolor[HTML]{9BB1AF}\textbf{$t_i$ (ms)} \\
\hline
RT-DETR & 85.0 & 82.8 & 89.1 & 53.2 & 3.7 \\
\hline
\textbf{CA RT-DETR (Ours)} & \textbf{88.8} & \textbf{86.4} & \textbf{91.6} & \textbf{56.1} & \textbf{4.1} \\
\hline
\end{tabular}%
}
\end{table}

These values demonstrate strong and consistent performance for detecting blue emergency lights under a wide range of conditions. The high mAP@\{0.5\} confirms the model’s ability to correctly localize lights in most cases, while the mAP@\{0.5:0.95\} indicates robustness even under stricter Intersection over Union (IoU) thresholds.

The model’s balanced precision and recall ensures a low false alarm rate while capturing most true instances, with only a slight, acceptable increase in inference time.

\subsection{Contextualized Results}
\label{subsec:contextualized}

\begin{figure*}[!ht]
    \centering
    \includegraphics[width=0.85\textwidth]{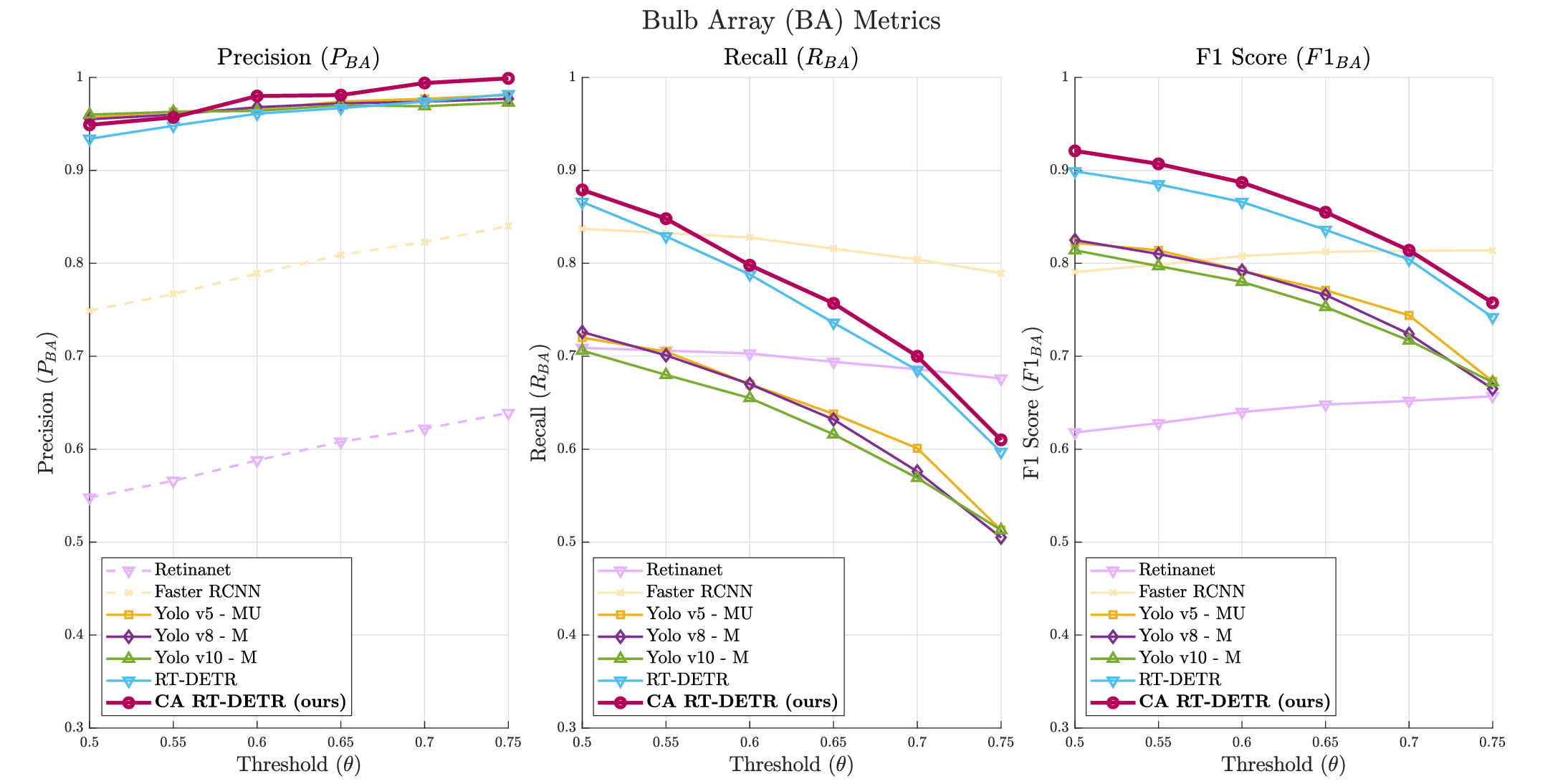}
    \caption{Comparison of metrics in Bulb Array grouping for different confidence values.}
    \label{fig:fig12}
\end{figure*}

\begin{figure*}[!ht]
    \centering
    \includegraphics[width=0.85\textwidth]{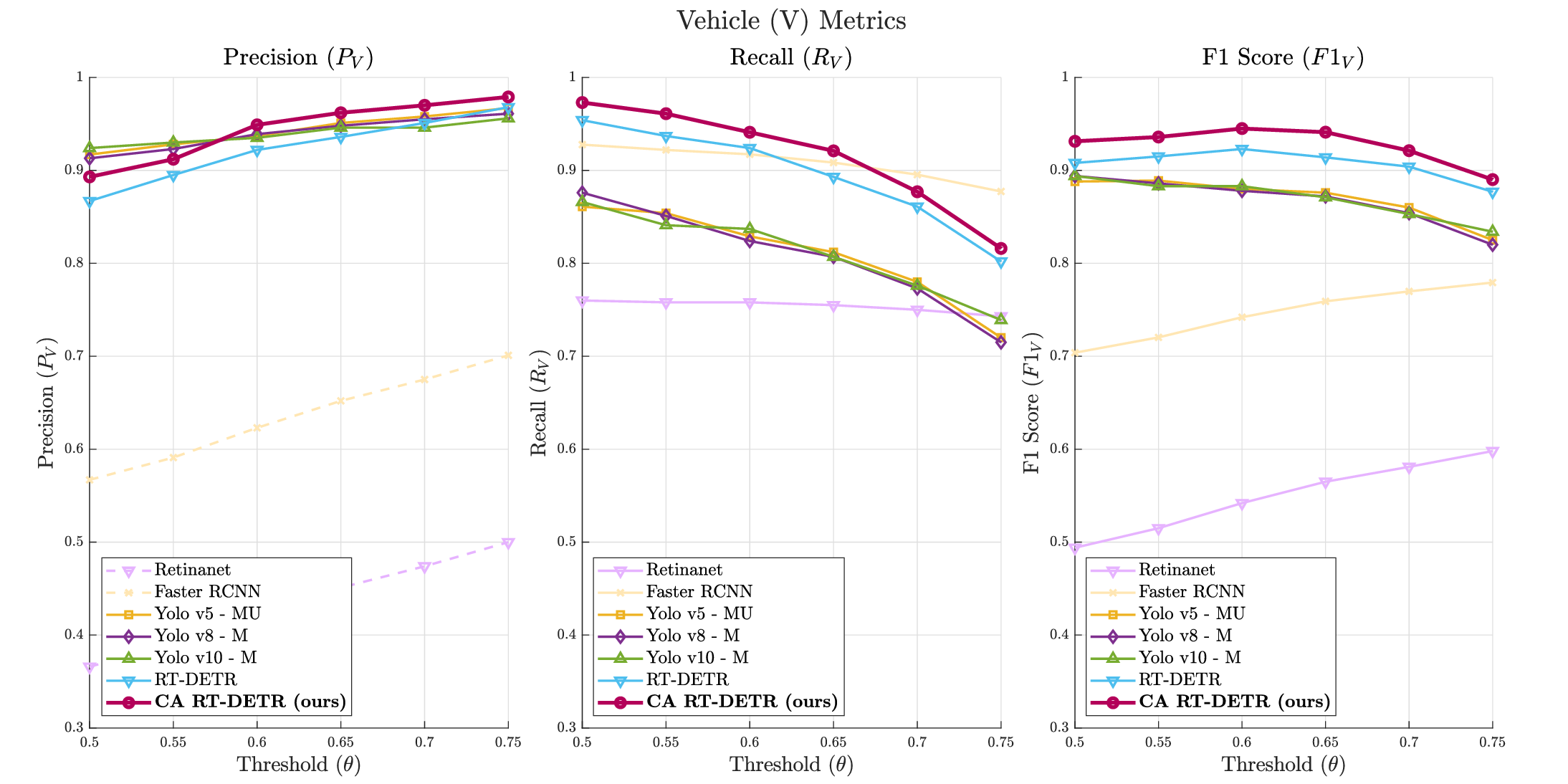}
    \caption{Comparison of metrics in Vehicle grouping for different confidence values.}
    \label{fig:fig13}
\end{figure*}

Figure~\ref{fig:fig12} and Figure~\ref{fig:fig13} illustrate the evolution of Precision (\textit{P}), Recall (\textit{R}), and F1 as the confidence threshold (\textit{Threshold}) for blue light detection is modified. Each model (RetinaNet, Faster R-CNN, YOLO v5--v8--v10, RT-DETR, and CA RT-DETR) is evaluated at two levels:
\begin{itemize}
    \item \textbf{Bulb Array (BA)}: Individual bulbs or groups of contiguous blue bulbs,
    \item \textbf{Vehicle (V)}: Detecting a single bulb suffices to consider the entire emergency vehicle as detected.
\end{itemize}
As the threshold increases, precision tends to grow (fewer false positives), while recall decreases (more false negatives).
\vspace{2mm}


First, all models increase their precision as the confidence threshold rises, at the cost of a decrease in recall, with varying rates of change across different models. For example, YOLO v5 -- MU starts with $P_{\mathrm{BA}} = 0.957$ and $R_{\mathrm{BA}} = 0.720$ ($\mathrm{F1}_{\mathrm{BA}} = 0.822$) at $\theta = 0.50$, but at $\theta = 0.75$, precision increases to $0.981$ while recall drops to $0.513$ ($\mathrm{F1}_{\mathrm{BA}} = 0.673$). Similarly, YOLO v8 -- M begins at $0.955$--$0.726$ ($\mathrm{F1}_{\mathrm{BA}} = 0.825$) and ends at $0.977$--$0.505$ ($\mathrm{F1}_{\mathrm{BA}} = 0.665$). YOLO v10 -- M shows equivalent figures: $P_{\mathrm{BA}}$ rises to $0.973$, but recall decreases to $0.513$, placing $\mathrm{F1}$ at $0.672$.

\begin{figure*}[!ht]
    \centering
    \includegraphics[width=0.85\textwidth]{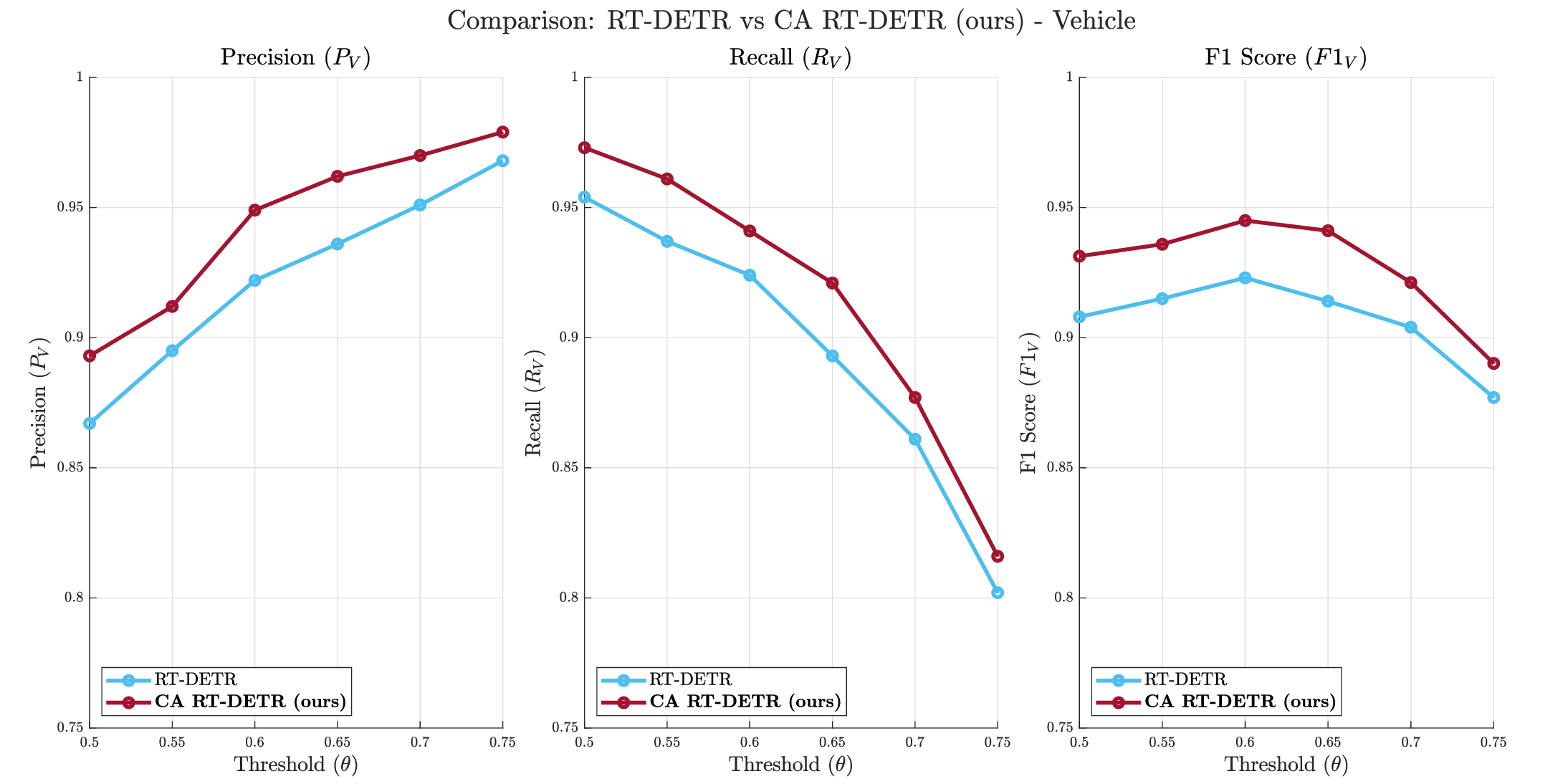}
    \caption{RT-DETR vs. CA RT-DETR vehicle metrics. The figure shows that incorporating the Color Attention module enhances vehicle detection performance by improving both precision and recall.}
    \label{fig:fig14}
\end{figure*}

Among the methods with lower volatility, Faster R-CNN maintains a relatively stable balance: with a low threshold ($\theta = 0.50$), it achieves $P_{\mathrm{BA}} = 0.749$ and $R_{\mathrm{BA}} = 0.837$ ($\mathrm{F1}_{\mathrm{BA}} = 0.791$), increasing precision to $\sim0.84$ at $\theta = 0.75$ while recall drops to $\sim0.79$ ($\mathrm{F1}_{\mathrm{BA}} = 0.813$). RetinaNet, on the other hand, starts with $P_{\mathrm{BA}} = 0.548$ and $R_{\mathrm{BA}} = 0.709$, rising to $P_{\mathrm{BA}} = 0.639$ and $R_{\mathrm{BA}} = 0.676$; its $\mathrm{F1}_{\mathrm{BA}}$ fluctuates between $\sim0.618$ and $0.657$, remaining below the rest. Although these models achieve acceptable recall values, their relatively low precision limits their reliability, leading to a higher rate of false positives. This compromises their applicability in scenarios where detection accuracy is critical, making them less suitable compared to more balanced alternatives.
\vspace{3mm} 

For its part, RT-DETR starts with $P_{\mathrm{BA}} = 0.934$, $R_{\mathrm{BA}} = 0.866$, and $\mathrm{F1}_{\mathrm{BA}} = 0.899$ at $\theta = 0.50$. In the high-threshold range ($\theta = 0.75$), precision rises to $0.982$, while recall drops to $0.597$, setting its $\mathrm{F1}_{\mathrm{BA}}$ at $0.742$. Finally, the enhanced version CA RT-DETR (\textit{ours}) gains even greater stability. At $0.50$, it achieves $P_{\mathrm{BA}} = 0.949$ and $R_{\mathrm{BA}} = 0.879$ ($\mathrm{F1}_{\mathrm{BA}} = 0.921$), and at $0.75$, it reaches $P_{\mathrm{BA}} = 0.999$ and $R_{\mathrm{BA}} = 0.61$ ($\mathrm{F1}_{\mathrm{BA}} = 0.756$). What makes the difference is the intermediate range ($\theta = 0.55$--$0.70$), where CA RT-DETR combines high precision ($>0.98$) with recall above $0.70$, surpassing the ``standard'' RT-DETR by several F1 points. In other words, the chromatic attention module allows the model to recognize bulbs (or light groups) in varied lighting conditions without triggering false positives at low and medium thresholds, nor losing too much recall when higher confidence is required.
\vspace{3mm} 

When considering the detection of entire vehicles (where detecting just one of their bulbs is sufficient), the initial recall of some models is high (Figure~\ref{fig:fig13}). For instance, Faster R-CNN starts at $R_V = 0.928$ and $P_V = 0.567$ ($\mathrm{F1}_V = 0.704$) at $\theta = 0.50$, increasing to $P_V = 0.701$ and $R_V = 0.877$ at $\theta = 0.75$ ($\mathrm{F1}_V = 0.779$). RetinaNet, on the other hand, ranks lower: it begins at $0.366$--$0.760$ ($\mathrm{F1}_V = 0.494$) and reaches $0.501$--$0.743$ ($\mathrm{F1}_V = 0.598$). Both models are ruled out due to their low precision, as their application would result in a high false positive rate---distracting the driver---that does not justify their stable recall.
Meanwhile, YOLO v5 -- MU achieves $P_V = 0.917$ and $R_V = 0.861$ ($\mathrm{F1}_V = 0.888$) at $\theta = 0.50$, but at $\theta = 0.75$, precision increases to $0.961$ while recall drops to $0.72$ ($\mathrm{F1}_V = 0.825$). Similar behavior is observed in YOLO v8 and v10, with precisions exceeding $0.95$ at $\theta = 0.75$ but recalls around $0.72$, resulting in F1 values around $0.83$. RT-DETR also experiences a recall adjustment but maintains $\mathrm{F1}_V = 0.877$ at $\theta = 0.75$, supported by $P_V = 0.968$ and $R_V = 0.802$.
\vspace{3mm} 

Once again, CA RT-DETR (\emph{ours}) proves to be the best option. At $\theta = 0.50$, it achieves $P_V = 0.893$, $R_V = 0.973$, and $\mathrm{F1}_V = 0.922$; and at $\theta = 0.75$, it reaches $P_V = 0.979$, $R_V = 0.816$ ($\mathrm{F1}_V = 0.89$), attaining $\mathrm{F1}_V$ values between $0.92$ and $0.95$ for intermediate thresholds. At a threshold of $0.6$, it achieves a precision of $94.9\%$ and a recall of $94.1\%$. This confirms that the color attention module enhances emergency vehicle detection more effectively (Fig.~\ref{fig:fig14}).
\vspace{3mm} 

\begin{figure}[!ht]
    \centering
    \subfloat[]{%
        \includegraphics[trim=110 20 5 20,clip,width=0.5\columnwidth]{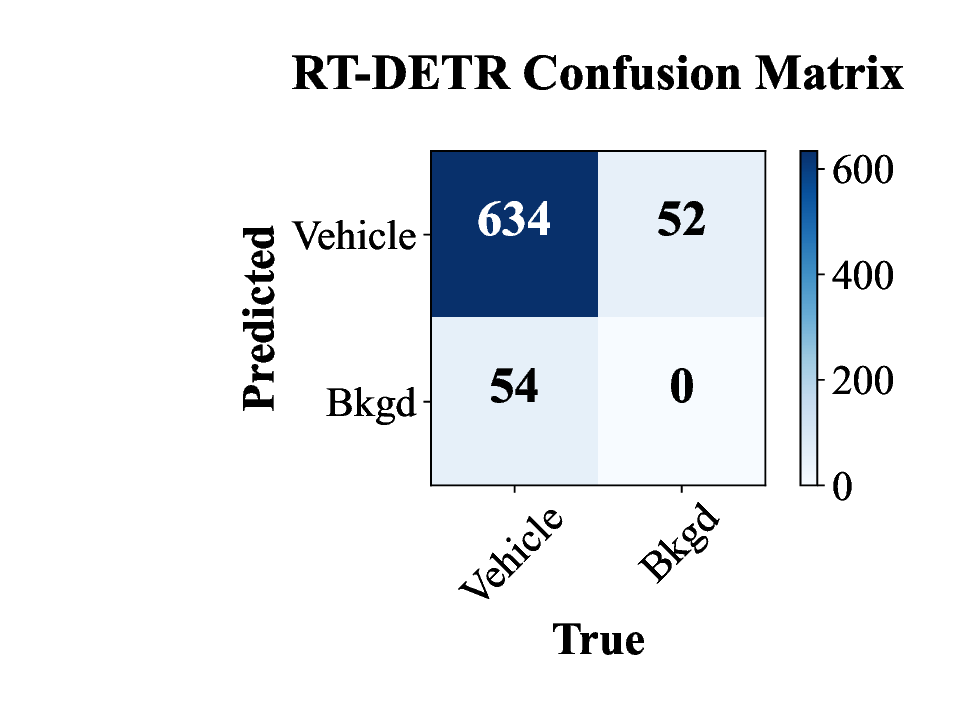}%
        \label{fig:RT}%
    }
    \hfill
    \subfloat[]{%
        \includegraphics[trim=110 20 5 20,clip,width=0.5\columnwidth]{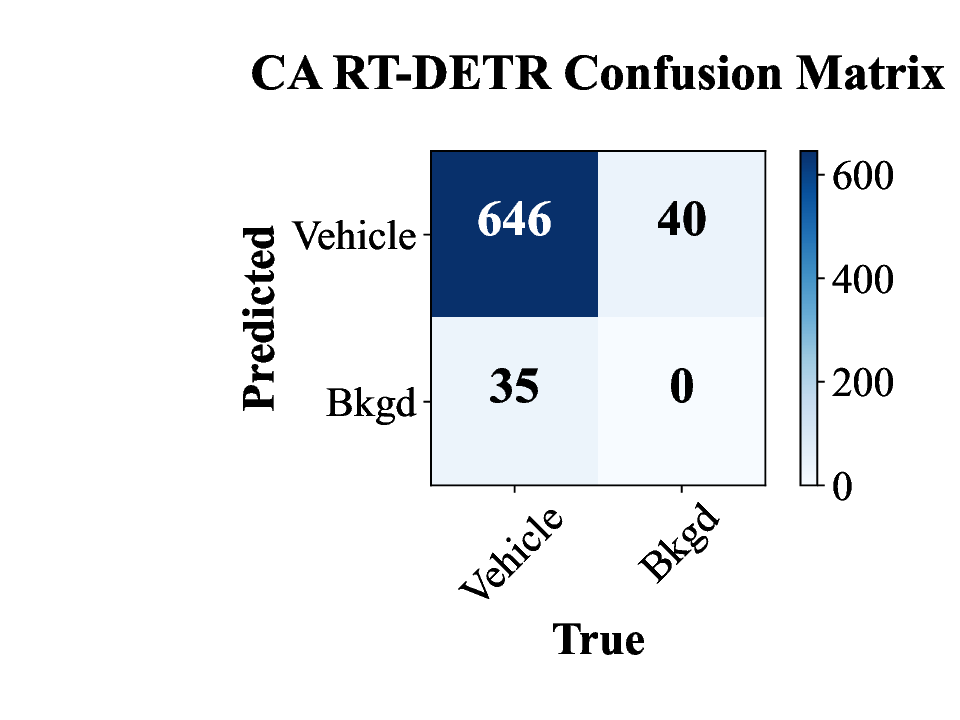}%
        \label{fig:CA}%
    }
    \caption{Confusion matrices for the RT-DETR (\ref{fig:RT}) and CA RT-DETR (ours) (\ref{fig:CA}) models. The CA RT-DETR model outperforms the baseline, showing a 1.9\% increase in true positives, a 35.2\% decrease in false negatives, and a 23.1\% reduction in false positives.}
    \label{fig:mat}
\end{figure}

Thus, CA RT-DETR establishes itself as the best overall solution for recognizing priority blue lights, both at the bulb array and vehicle levels, and it was the model selected for implementation in the 360\degree{} detection system (Table~\ref{tab:table3}).
 \\ \\
\begin{table}[!ht]
\centering
\caption{Best F1 Scores with Corresponding Precision and Recall for Bulb Array and Vehicle Detection.}
\label{tab:table3}
\resizebox{\columnwidth}{!}{%
\begin{tabular}{|c|ccc|ccc|}
\hline
\rowcolor[HTML]{9BB1AF} 
\cellcolor[HTML]{C8DAD5} &
  \multicolumn{3}{c|}{\cellcolor[HTML]{9BB1AF}\textbf{Bulb Array}} &
  \multicolumn{3}{c|}{\cellcolor[HTML]{9BB1AF}\textbf{Vehicle}} \\ \cline{2-7} 
\rowcolor[HTML]{D6F1EA} 
\multirow{-2}{*}{\cellcolor[HTML]{C8DAD5}\textbf{Model}} &
  \textbf{Best $F1_{BA}$} &
  \textbf{$P_{BA}$} &
  \textbf{$R_{BA}$} &
  \textbf{Best $F1_V$} &
  \textbf{$P_V$} &
  \textbf{$R_V$} \\ \hline
Retinanet    & 65.7  & 63.9 & 67.6 & 49.4  & 50.0 & 76.0 \\ \cline{1-1}
Faster R-CNN & 81.39 & 84.0 & 78.9 & 77.93 & 70.1 & 87.7 \\ \cline{1-1}
Yolo v5 - MU & 82.2  & 95.7 & 72.0 & 88.9  & 92.8 & 85.4 \\ \cline{1-1}
Yolo v8 - M  & 82.5  & 95.5 & 72.6 & 89.4  & 91.3 & 87.6 \\ \cline{1-1}
Yolo v10 - M & 81.4  & 96.0 & 70.6 & 89.4  & 92.4 & 86.6 \\ \cline{1-1}
RT-DETR      & 89.9  & 93.4 & 86.6 & 92.3  & 92.2 & 92.4 \\ \cline{1-1}
\textbf{\begin{tabular}[c]{@{}c@{}}CA RT-DETR \\ (ours)\end{tabular}} &
  \textbf{92.1} &
  \textbf{94.9} &
  \textbf{87.9} &
  \textbf{94.5} &
  \textbf{94.9} &
  \textbf{94.1} \\ \hline
\end{tabular}%
}
\end{table}
\newpage In Figure \ref{fig:mat}, it can be observed that our model achieves 1.9\% increase in true positives, a 35.2\% reduction in false negatives and a 23.1\% reduction in false positives.

\subsection{Field Test Results}
\label{subsec:fieldTests}
\begin{figure}[!ht]
    \centering
    \includegraphics[width=\columnwidth]{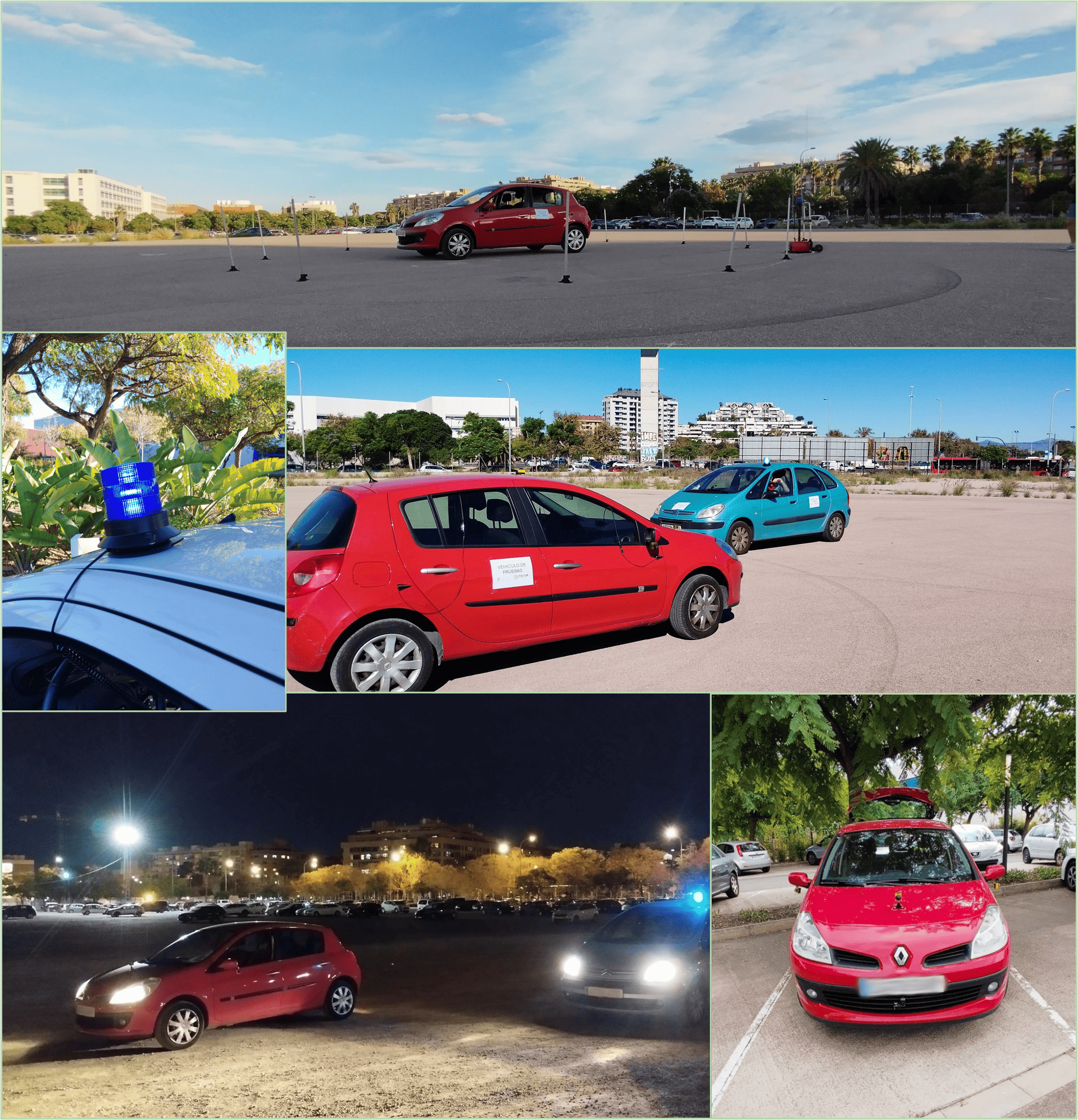}
    \caption{Field test images capturing both static and dynamic scenarios, using instrumented vehicles to measure and simulate real-world emergency vehicle presence.}
    \label{fig:fig15}
\end{figure}

To validate the performance of the proposed system in a setting close to real-world driving, field tests were conducted using two passenger vehicles: the first equipped with four fisheye cameras and the second with an active blue emergency light (Fig.~\ref{fig:fig15}). By synchronizing the positions of both vehicles (whose coordinates were precisely known), various approach and distance scenarios were recreated under both daytime and nighttime conditions. The primary objective was to quantitatively assess the precision, recall, and error in the estimation of the azimuthal detection angle.

The results obtained from testing the proposed system in two different scenarios are described below. Field tests were conducted in two distinct setups:
\begin{itemize}
    \item \textit{Static tests:} The vehicle equipped with cameras remains stationary while the car with the emergency light is positioned at different angles and distances.
    \item \textit{Dynamic tests:} Both vehicles in motion to simulate real traffic conditions.
\end{itemize}

In all trials, a single emergency light of relatively small dimensions and lower power than those typically used on standard emergency vehicles was employed (similar to those installed on secret police cars or medium-sized ambulances), making the detection challenge more demanding. Even so, it is demonstrated that the CA RT-DETR model, together with the 360\degree{} multi-camera configuration, achieves high accuracy rates, even under varied lighting conditions and distances, as will be shown below.

\subsubsection{Static Tests}
\label{subsubsec:staticTests}
In this experiment, the car equipped with the multi-camera system remained stationary, while the vehicle with the emergency light was placed at different azimuthal angles in 15\degree{} increments (0\degree, 15\degree, 30\degree, \ldots, 345\degree) and at increasing distances from the vehicle with cameras. For each combination of angle and distance, multiple frames (videos) were captured to analyze:
\begin{itemize}
    \item \textbf{Detection Rate (DR)}: The percentage of frames in which the blue light is correctly detected at its actual position.
    \item \textbf{Azimuthal error} ($E_{\varphi}$): The difference between the angle estimated by the system and the actual angle at which the emergency light vehicle was positioned (in degrees).
\end{itemize}

Table~\ref{tab:staticTests} shows the average values of DR and $E_{\varphi}$ considering different distances under both daytime and nighttime conditions. For each distance, measurements were taken at all angles, and an average azimuthal error value was obtained.

\begin{table}[!ht]
\centering
\caption{System performance in static tests at different distances, detailing whether detection was achieved (Det.), the detection rate (DR), and the average azimuthal ($\mathbf{E_{\varphi}(^\circ)}$) under both daytime and nighttime conditions.}
\label{tab:staticTests}
\resizebox{\columnwidth}{!}{%
\begin{tabular}{|c|ccc|ccc|}
\hline
\rowcolor[HTML]{9BB1AF} 
\cellcolor[HTML]{C8DAD5}                               & \multicolumn{3}{c|}{\cellcolor[HTML]{9BB1AF}Day}     & \multicolumn{3}{c|}{\cellcolor[HTML]{9BB1AF}Night}   \\
\rowcolor[HTML]{C8DAD5} 
\multirow{-2}{*}{\cellcolor[HTML]{C8DAD5}\begin{tabular}[c]{@{}c@{}}Distance \\ (m)\end{tabular}} &
  Det. &
  DR (\%) &
  $\mathbf{E_{\varphi}(^\circ)}$ &
  Det. &
  DR (\%) &
  $\mathbf{E_{\varphi}(^\circ)}$ \\ \hline
10 & \checkmark & 100.0 & \(\sim1.2\) & \checkmark & 100.0 & \(\sim1.1\) \\
20 & \checkmark & 100.0 & \(\sim1.9\) & \checkmark & 100.0 & \(\sim1.2\) \\
30 & \checkmark & 98.3  & \(\sim2.0\) & \checkmark & 100.0 & \(\sim2.4\) \\
40 & \checkmark & 63.4  & \(\sim2.2\) & \checkmark & 99.2  & \(\sim2.5\) \\
50 & \checkmark & 22.9  & \(\sim2.2\) & \checkmark & 93.1  & \(\sim2.6\) \\
60 & \checkmark & 6.4   & \(\sim2.4\) & \checkmark & 59.3  & \(\sim2.9\) \\
70 & \(\times\)                & 0.0   & --          & \checkmark & 31.2  & \(\sim3.0\) \\
80 & \(\times\)                & --    & --          & \checkmark & 1.4   & \(\sim2.4\) \\
90 & \(\times\)                & --    & --          & \(\times\)                & 0.0   & --          \\ \hline
\end{tabular}%
}
\end{table}

\begin{figure}[!ht]
    \centering
    \includegraphics[width=\columnwidth]{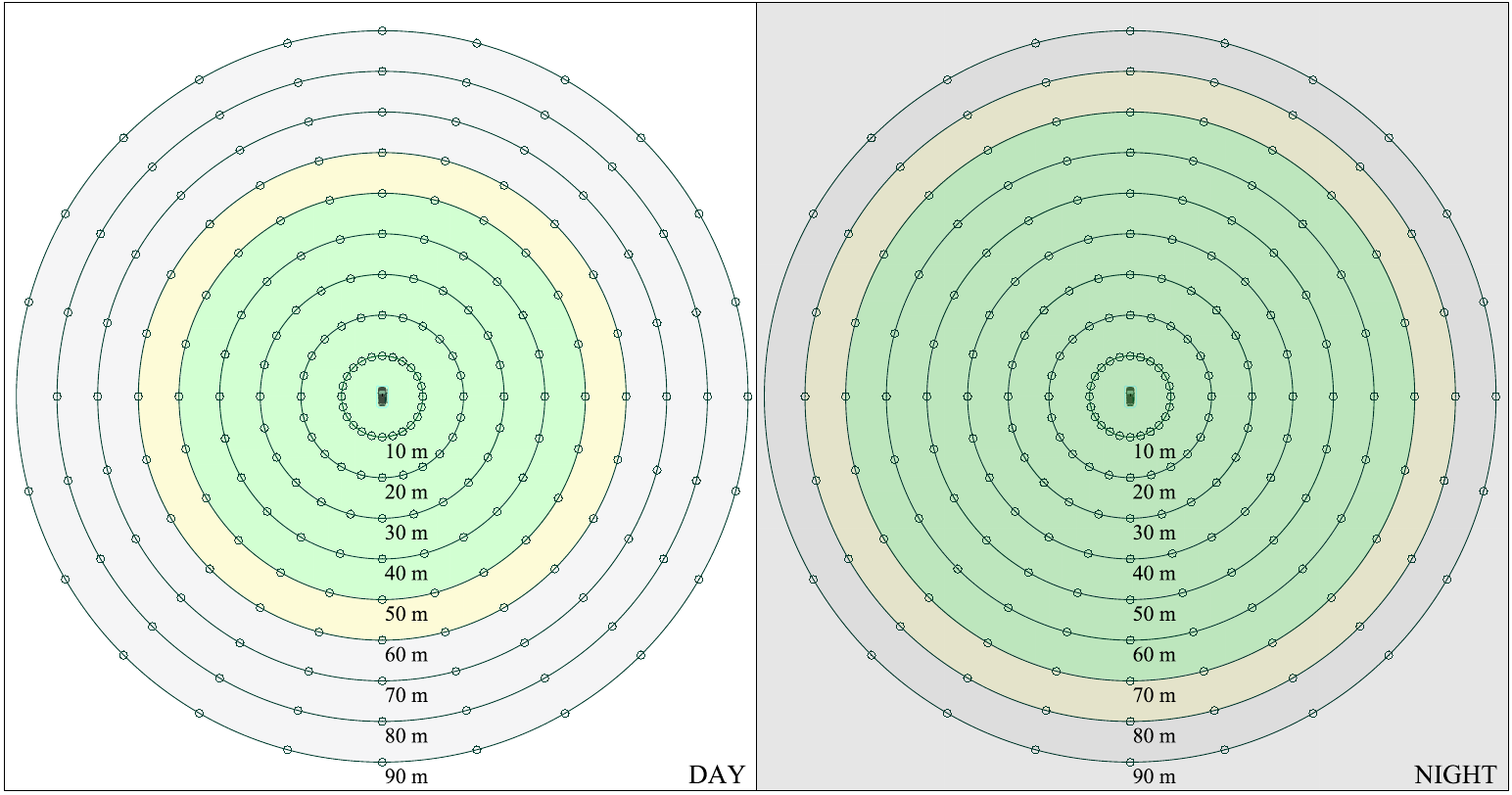}
    \caption{System detection ranges in static tests under daytime and nighttime conditions, illustrating how far the system can reliably detect active emergency lights.}
    \label{fig:fig16}
\end{figure}

In general terms, it can be observed that during the day, detection is practically perfect up to 20–30 meters, with an azimuthal error remaining below 2.0\degree{} on average. It should be noted that these tests were conducted using a relatively small emergency light; therefore, in normal conditions with larger, more powerful lighting structures, the detection performance is expected to be even better. Beyond 40 meters, the detection rate drops more sharply due to the smaller apparent size of the light in the image, although some frames still capture the blue flash. Detection is lost beyond 60 meters, as the emergency light becomes too small to be reliably recognized (Fig.~\ref{fig:fig16}).

At night, the contrast of the light enhances detection, achieving nearly 100\% accuracy up to 40--50 meters. Even beyond 50--60 meters, partial detections are still obtained. While not present in every frame, these intermittent detections still confirm the presence of the emergency light.

It is important to emphasize that, although 100\% detection is not achieved in all frames at greater distances, the system continues to register occasional detections that are sufficient to alert the presence of a priority vehicle. In practice, having just a few valid frames within a given time interval would be enough to fulfill the objective of identifying the blue light and notifying the driver or the assistance system of the approach of an emergency vehicle.

\subsubsection{Dynamic Tests}
\label{subsubsec:dynamicTests}
To evaluate the reliability of the system in more realistic traffic situations, several movement patterns were designed to simulate different approaches of an emergency vehicle: frontal, rear, lateral, complete turns around the car, and semicircular trajectories. These paths cover the most common cases of urban and highway driving, where an emergency vehicle may appear from any angle. The goal was to replicate everyday maneuvers such as overtaking, crossing, or direct approach.

Table~\ref{tab:dynamicTests} presents the Detection Rate (DR) in both daytime and nighttime conditions, along with the distance ranges at which detection was consistently maintained throughout the passes. Each scenario was repeated between 10 and 20 times to obtain representative values, confirming that the light was detected most of the time, with only brief intermittent losses when the emergency vehicle moved significantly farther away.

\begin{table}[!ht]
\centering
\caption{System performance in dynamic tests simulating various emergency vehicle approaches. The table reports the approximate speed, detection rate as a percentage, and the distance range between the measurement vehicle and the emergency vehicle in which the test was conducted, under both daytime and nighttime conditions.}
\label{tab:dynamicTests}
\resizebox{\columnwidth}{!}{%
\begin{tabular}{|l|c|cc|cc|}
\hline
\rowcolor[HTML]{9BB1AF} 
\cellcolor[HTML]{C8DAD5} &
  \cellcolor[HTML]{9BB1AF} &
  \multicolumn{2}{c|}{\cellcolor[HTML]{9BB1AF}\textbf{Day}} &
  \multicolumn{2}{c|}{\cellcolor[HTML]{9BB1AF}\textbf{Night}} \\ \cline{3-6} 
\rowcolor[HTML]{D6F1EA} 
\multirow{-2}{*}{\cellcolor[HTML]{C8DAD5}\textbf{Scenario}} &
  \multirow{-2}{*}{\cellcolor[HTML]{9BB1AF}\textbf{\begin{tabular}[c]{@{}c@{}}Approx Vel. \\ (km/h)\end{tabular}}} &
  \textbf{DR (\%)} &
  \textbf{Dist. (m)} &
  \textbf{DR (\%)} &
  \textbf{Dist. (m)} \\ \hline
\begin{tabular}[c]{@{}l@{}}Frontal \\ approach\end{tabular}          & 30 & 92.5 & 2-50   & 99.8  & 2-50   \\ \hline
\begin{tabular}[c]{@{}l@{}}Rear \\ approach\end{tabular}             & 30 & 91.2 & 2-50   & 99.7  & 2-50   \\ \hline
\begin{tabular}[c]{@{}l@{}}Lateral \\ approach\end{tabular}          & 50 & 93.7 & 2-50   & 99.9  & 2-50   \\ \hline
\begin{tabular}[c]{@{}l@{}}360º loop \\ around car\end{tabular}      & 20 & 99.6 & 20-30  & 100.0 & 20-30  \\ \hline
\begin{tabular}[c]{@{}l@{}}Semicircular \\ route (180º)\end{tabular} & 40 & 90.2 & 50-1.5 & 99.2  & 50-1.5 \\ \hline
\end{tabular}%
}
\end{table}

In general terms, the Detection Rate exceeds 91\% during the day and 99\% at night. The range of distances at which the light is consistently perceived varies between 0--40\,meters in daytime conditions, extending up to 60\,meters at night due to the increased contrast. Even when brief intervals occur where no detection is made in individual frames, the system quickly recovers the signal in successive captures. In practice, this ensures that at least one valid frame is always available to confirm the presence of a priority vehicle.

This behavior is crucial for a driver assistance system, as it guarantees timely alerts to the driver, even in the case of rapid maneuvers or unexpected trajectories of the emergency vehicle.

\section{Conclusions}
\label{sec:conclusions}

In this work, an integrated emergency vehicle detection system has been presented, based on the identification of active blue lights through a 360\degree{} multi-camera configuration and an improved deep learning model (CA~RT-DETR). The premise is that detecting any active blue light from an emergency vehicle is sufficient to consider it identified, reducing the complexity of the problem and prioritizing a high detection rate in critical scenarios.

To train and validate the model, the Active Blue Light Dataset (ABLDataset) has been created---a novel resource that compiles thousands of images and annotations of emergency blue lights under diverse conditions. This approach covers multiple environments, distances, and vehicle types, ensuring broad variability in the dataset. With the aim of fostering research advancements in the detection of priority vehicles, the ABLDataset is being shared with the scientific community, enabling other researchers to benefit from this resource and develop new approaches or refine existing ones.

The results of the experimental evaluation show that incorporating a chromatic attention module into the RT-DETR architecture significantly improves the detection of blue lights compared to reference models such as YOLO (v5, v8, v10), RetinaNet, and Faster R-CNN. This modification enhances the model's response to variations in color tones and light intensities encountered in real traffic environments. This gain in robustness proves particularly useful at medium and long distances, as well as in low-visibility or high-illumination scenarios.

Furthermore, a 360\degree{} multi-camera prototype based on fisheye lenses has been developed and tested. Its design and calibration allow for complete perimeter coverage of the vehicle and the estimation of the angular position of the detection. This subsystem contributes to road safety by alerting the driver to the presence and direction of emergency vehicles, all with low inference times and minimal false-negative rates. Field tests conducted at various distance ranges and under daytime and nighttime conditions have demonstrated that the proposed solution maintains high precision values, with an azimuthal angle estimation error generally below $3^\circ$.

In summary, the system presented combines a carefully designed data acquisition methodology, a training strategy that maximizes blue light detection, and a multi-camera configuration for 360\degree{} coverage. This approach not only enables identification of a priority vehicle but also provides the driver with a precise spatial reference of the emergency signal's origin. The release of ABLDataset to the research community aims to accelerate the adoption and improvement of these solutions, promoting the integration of emergency vehicle detection in Advanced Driver Assistance Systems and, potentially, in next-generation autonomous vehicles. Ultimately, this contributes to enhancing road safety and reducing reaction times in critical scenarios, leading to more efficient and safer traffic for all road users.

\bibliographystyle{IEEEtran}
\bibliography{refs}

@misc{JuanEmergencias2024juanemergenciasYouTube,
    title = {{@juanemergencias - YouTube}},
    year = {2024},
    author = {{Juan Emergencias}},
    url = {https://www.youtube.com/@juanemergencias}
}

@misc{UrgencesGeneve2024UrgencesGeneveYouTube,
    title = {{@UrgencesGeneve - YouTube}},
    year = {2024},
    author = {{Urgences Geneve}},
    url = {https://www.youtube.com/@UrgencesGeneve}
}

@article{Kannala2006ALenses,
    title = {{A generic camera model and calibration method for conventional, wide-angle, and fish-eye lenses}},
    year = {2006},
    journal = {IEEE Transactions on Pattern Analysis and Machine Intelligence},
    author = {Kannala, Juho and Brandt, Sami S.},
    number = {8},
    volume = {28},
    doi = {10.1109/TPAMI.2006.153},
    issn = {01628828}
}

@article{Li2022AChina,
    title = {{A Survey of ADAS Perceptions With Development in China}},
    year = {2022},
    journal = {IEEE Transactions on Intelligent Transportation Systems},
    author = {Li, Xinran and Lin, Kuo Yi and Meng, Min and Li, Xiuxian and Li, Li and Hong, Yiguang and Chen, Jie},
    number = {9},
    volume = {23},
    doi = {10.1109/TITS.2022.3149763},
    issn = {15580016}
}

@article{Shorten2019ALearning,
    title = {{A survey on Image Data Augmentation for Deep Learning}},
    year = {2019},
    journal = {Journal of Big Data},
    author = {Shorten, Connor and Khoshgoftaar, Taghi M.},
    number = {1},
    volume = {6},
    doi = {10.1186/s40537-019-0197-0},
    issn = {21961115}
}

@article{Peng2023ANetworking,
    title = {{A Survey on In-Vehicle Time-Sensitive Networking}},
    year = {2023},
    journal = {IEEE Internet of Things Journal},
    author = {Peng, Yifei and Shi, Boxin and Jiang, Tigang and Tu, Xiaodong and Xu, Du and Hua, Kun},
    number = {16},
    volume = {10},
    doi = {10.1109/JIOT.2023.3264909},
    issn = {23274662}
}

@article{Ashir2022ADetection,
    title = {{A Transfer-Learning-Based Approach for Emergency Vehicle Detection}},
    year = {2022},
    journal = {Eurasian Journal of Science and Engineering},
    author = {Ashir, Abubakar},
    number = {1},
    volume = {8},
    doi = {10.23918/eajse.v8i1p75},
    issn = {24145629}
}

@inproceedings{Mittal2022AcousticModels,
    title = {{Acoustic Based Emergency Vehicle Detection Using Ensemble of deep Learning Models}},
    year = {2022},
    booktitle = {Procedia Computer Science},
    author = {Mittal, Usha and Chawla, Priyanka},
    volume = {218},
    doi = {10.1016/j.procs.2023.01.005},
    issn = {18770509}
}

@article{Tran2020Acoustic-BasedNetworks,
    title = {{Acoustic-Based Emergency Vehicle Detection Using Convolutional Neural Networks}},
    year = {2020},
    journal = {IEEE Access},
    author = {Tran, Van Thuan and Tsai, Wei Ho},
    volume = {8},
    doi = {10.1109/ACCESS.2020.2988986},
    issn = {21693536}
}

@article{Wu2023ADD:Driving,
    title = {{ADD: An automatic desensitization fisheye dataset for autonomous driving}},
    year = {2023},
    journal = {Engineering Applications of Artificial Intelligence},
    author = {Wu, Zizhang and Chen, Xinyuan and Wei, Hongyang and Song, Fan and Xu, Tianhao},
    pages = {106766},
    volume = {126},
    url = {https://www.sciencedirect.com/science/article/pii/S0952197623009508},
    doi = {https://doi.org/10.1016/j.engappai.2023.106766},
    issn = {0952-1976}
}

@article{Shatnawi2024AnGAN,
    title = {{An Enhanced Model for Detecting and Classifying Emergency Vehicles Using a Generative Adversarial Network (GAN)}},
    year = {2024},
    journal = {Vehicles},
    author = {Shatnawi, Mo’ath and Bani Younes, Maram},
    number = {3},
    pages = {1114--1139},
    volume = {6},
    doi = {10.3390/vehicles6030053},
    issn = {2624-8921}
}

@article{Lee2023AnalysisVehicles,
    title = {{Analysis of real-time operating systems’ file systems: Built-in cameras from vehicles}},
    year = {2023},
    journal = {Forensic Science International: Digital Investigation},
    author = {Lee, Jung Hwan and Hyeon, Bum Su and Jeon, Oc Yeub and Park, Nam In},
    volume = {44},
    doi = {10.1016/j.fsidi.2023.301500},
    issn = {26662817}
}

@article{Tran2021Audio-VisionDetection,
    title = {{Audio-Vision Emergency Vehicle Detection}},
    year = {2021},
    journal = {IEEE Sensors Journal},
    author = {Tran, Van Thuan and Tsai, Wei Ho},
    number = {24},
    volume = {21},
    doi = {10.1109/JSEN.2021.3127893},
    issn = {15581748}
}

@inproceedings{Yu2020BDD100K:Learning,
    title = {{BDD100K: A Diverse Driving Dataset for Heterogeneous Multitask Learning}},
    year = {2020},
    booktitle = {Proceedings of the IEEE Computer Society Conference on Computer Vision and Pattern Recognition},
    author = {Yu, Fisher and Chen, Haofeng and Wang, Xin and Xian, Wenqi and Chen, Yingying and Liu, Fangchen and Madhavan, Vashisht and Darrell, Trevor},
    doi = {10.1109/CVPR42600.2020.00271},
    issn = {10636919}
}

@article{BROWNDC1971Close-Calibration,
    title = {{Close- range camera calibration}},
    year = {1971},
    journal = {Photogramm Eng},
    author = {{BROWN DC}},
    number = {8},
    volume = {37}
}

@misc{Inc.2018CreateDocumentation,
    title = {{Create ML documentation}},
    year = {2018},
    author = {Inc., Apple},
    howpublished = {https://developer.apple.com/documentation/createml}
}

@article{Hossain2023DataModes,
    title = {{Data mining approach to explore emergency vehicle crash patterns: A comparative study of crash severity in emergency and non-emergency response modes}},
    year = {2023},
    journal = {Accident Analysis and Prevention},
    author = {Hossain, Md Mahmud and Zhou, Huaguo and Das, Subasish},
    pages = {107217},
    volume = {191},
    doi = {10.1016/j.aap.2023.107217}
}

@article{Liu2020DeepSurvey,
    title = {{Deep Learning for Generic Object Detection: A Survey}},
    year = {2020},
    journal = {International Journal of Computer Vision},
    author = {Liu, Li and Ouyang, Wanli and Wang, Xiaogang and Fieguth, Paul and Chen, Jie and Liu, Xinwang and Pietik{\"{a}}inen, Matti},
    number = {2},
    volume = {128},
    doi = {10.1007/s11263-019-01247-4},
    issn = {15731405}
}

@article{StojmenovaPececnik2023DesignVehicles,
    title = {{Design of head-up display interfaces for automated vehicles}},
    year = {2023},
    journal = {International Journal of Human Computer Studies},
    author = {Stojmenova Pe{\v{c}}e{\v{c}}nik, Kristina and Toma{\v{z}}i{\v{c}}, Sašo and Sodnik, Jaka},
    volume = {177},
    doi = {10.1016/j.ijhcs.2023.103060},
    issn = {10959300}
}

@article{Parineh2023DetectingSignals,
    title = {{Detecting emergency vehicles With 1D-CNN using fourier processed audio signals}},
    year = {2023},
    journal = {Measurement: Journal of the International Measurement Confederation},
    author = {Parineh, Hossein and Sarvi, Majid and Bagloee, Saeed Asadi},
    volume = {223},
    doi = {10.1016/j.measurement.2023.113784},
    issn = {02632241}
}

@misc{Zhao2024DETRsDetection,
    title = {{DETRs Beat YOLOs on Real-time Object Detection}},
    year = {2024},
    author = {Zhao, Yian and Lv, Wenyu and Xu, Shangliang and Wei, Jinman and Wang, Guanzhong and Dang, Qingqing and Liu, Yi and Chen, Jie},
    url = {https://arxiv.org/abs/2304.08069},
    arxivId = {2304.08069}
}

@article{Lowe2004DistinctiveKeypoints,
    title = {{Distinctive image features from scale-invariant keypoints}},
    year = {2004},
    journal = {International Journal of Computer Vision},
    author = {Lowe, David G.},
    number = {2},
    volume = {60},
    doi = {10.1023/B:VISI.0000029664.99615.94},
    issn = {09205691}
}

@article{Weibull2025DriversIntroduction,
    title = {{Driver's gaze behavior when approached by an emergency vehicle – The effects of in-car warnings and system introduction}},
    year = {2025},
    journal = {Transportation Research Part F: Psychology and Behaviour},
    author = {Weibull, Kajsa and Lidestam, Björn and Prytz, Erik},
    pages = {137--146},
    volume = {109},
    doi = {10.1016/j.trf.2024.12.010}
}

@article{Zohaib2024EnhancingFusion,
    title = {{Enhancing Emergency Vehicle Detection: A Deep Learning Approach with Multimodal Fusion}},
    year = {2024},
    journal = {Mathematics},
    author = {Zohaib, Muhammad and Asim, Muhammad and ELAffendi, Mohammed},
    number = {10},
    volume = {12},
    doi = {10.3390/math12101514},
    issn = {2227-7390}
}

@article{Banchero2025EnhancingSound,
    title = {{Enhancing Road Safety with AI-Powered System for Effective Detection and Localization of Emergency Vehicles by Sound}},
    year = {2025},
    journal = {Sensors},
    author = {Banchero, Lucas and Vacalebri-Lloret, Francisco and Mossi, Jose M and Lopez, Jose J},
    number = {3},
    volume = {25},
    doi = {10.3390/s25030793},
    issn = {1424-8220}
}

@inproceedings{Knorr2014ExtrinsicView,
    title = {{Extrinsic calibration of a fisheye multi-camera setup using overlapping fields of view}},
    year = {2014},
    booktitle = {IEEE Intelligent Vehicles Symposium, Proceedings},
    author = {Knorr, Moritz and Esparza, Jose and Niehsen, Wolfgang and Stiller, Christoph},
    doi = {10.1109/IVS.2014.6856403}
}

@inproceedings{Ren2015FasterNetworks,
    title = {{Faster R-CNN: Towards real-time object detection with region proposal networks}},
    year = {2015},
    booktitle = {Advances in Neural Information Processing Systems},
    author = {Ren, Shaoqing and He, Kaiming and Girshick, Ross and Sun, Jian},
    volume = {2015-January},
    issn = {10495258}
}

@article{ADASBiro2023FisheyeDatasetDataset,
    title = {{FisheyeDataset Dataset}},
    year = {2023},
    journal = {Roboflow Universe},
    author = {{ADASBiro}},
    month = {10}
}

@article{Li2020FisheyeDet:Images,
    title = {{FisheyeDet: A self-study and contour-based object detector in fisheye images}},
    year = {2020},
    journal = {IEEE Access},
    author = {Li, Tangwei and Tong, Guanjun and Tang, Hongying and Li, Baoqing and Chen, Bo},
    volume = {8},
    doi = {10.1109/ACCESS.2020.2987868},
    issn = {21693536}
}

@inproceedings{Lin2017FocalDetection,
    title = {{Focal Loss for Dense Object Detection}},
    year = {2017},
    booktitle = {Proceedings of the IEEE International Conference on Computer Vision},
    author = {Lin, Tsung Yi and Goyal, Priya and Girshick, Ross and He, Kaiming and Dollar, Piotr},
    volume = {2017-October},
    doi = {10.1109/ICCV.2017.324},
    issn = {15505499}
}

@inproceedings{Bullough2019ImpactsGlare,
    title = {{Impacts of Flashing Emergency Lights and Vehicle-Mounted Illumination on Driver Visibility and Glare}},
    year = {2019},
    booktitle = {SAE Technical Paper 2019-01-0847},
    author = {Bullough, John and Skinner, Nicholas and Rea, Mark},
    month = {12},
    pages = {},
    doi = {10.4271/2019-01-0847}
}

@inproceedings{Kaushik2020LeveragingAnalysis,
    title = {{Leveraging Computer Vision for Emergency Vehicle Detection-Implementation and Analysis}},
    year = {2020},
    booktitle = {2020 11th International Conference on Computing, Communication and Networking Technologies, ICCCNT 2020},
    author = {Kaushik, S. and Raman, Abhishek and Rajeswara Rao, K. V.S.},
    doi = {10.1109/ICCCNT49239.2020.9225331}
}

@article{Yu2024MCG-RTDETR:Imagery,
    title = {{MCG-RTDETR: Multi-Convolution and Context-Guided Network with Cascaded Group Attention for Object Detection in Unmanned Aerial Vehicle Imagery}},
    year = {2024},
    journal = {Remote Sensing},
    author = {Yu, Chushi and Shin, Yoan},
    number = {17},
    volume = {16},
    doi = {10.3390/rs16173169},
    issn = {2072-4292}
}

@article{Joshi2024Multi-ModalVehicles,
    title = {{Multi-Modal Information Fusion for Localization of Emergency Vehicles}},
    year = {2024},
    journal = {International Journal of Image and Graphics},
    author = {Joshi, Aruna Kumar and Kulkarni, Shrinivasrao B.},
    doi = {10.1142/S0219467825500500},
    issn = {17936756}
}

@article{GladiensyahBihanda2024Multi-VehicleByteTrack,
    title = {{Multi-Vehicle Tracking and Counting Framework in Average Daily Traffic Survey Using RT-DETR and ByteTrack}},
    year = {2024},
    journal = {IEEE Access},
    author = {Gladiensyah Bihanda, Yusuf and Fatichah, Chastine and Yuniarti, Anny},
    number = {},
    pages = {121723--121737},
    volume = {12},
    doi = {10.1109/ACCESS.2024.3453249}
}

@article{Dazlee2022ObjectYOLO,
    title = {{Object Detection for Autonomous Vehicles with Sensor-based Technology Using YOLO}},
    year = {2022},
    journal = {International Journal of Intelligent Systems and Applications in Engineering},
    author = {Dazlee, Nurin Miza Afiqah Andrie and Khalil, Syamimi Abdul and Abdul-Rahman, Shuzlina and Mutalib, Sofianita},
    number = {1},
    volume = {10},
    doi = {10.18201/ijisae.2022.276},
    issn = {21476799}
}

@article{Zou2023ObjectSurvey,
    title = {{Object Detection in 20 Years: A Survey}},
    year = {2023},
    journal = {Proceedings of the IEEE},
    author = {Zou, Zhengxia and Chen, Keyan and Shi, Zhenwei and Guo, Yuhong and Ye, Jieping},
    number = {3},
    volume = {111},
    doi = {10.1109/JPROC.2023.3238524},
    issn = {15582256}
}

@misc{Zhao2019ObjectReview,
    title = {{Object Detection with Deep Learning: A Review}},
    year = {2019},
    booktitle = {IEEE Transactions on Neural Networks and Learning Systems},
    author = {Zhao, Zhong Qiu and Zheng, Peng and Xu, Shou Tao and Wu, Xindong},
    number = {11},
    volume = {30},
    doi = {10.1109/TNNLS.2018.2876865},
    issn = {21622388}
}

@article{Kumar2021OmniDet:Driving,
    title = {{OmniDet: Surround View Cameras Based Multi-Task Visual Perception Network for Autonomous Driving}},
    year = {2021},
    journal = {IEEE Robotics and Automation Letters},
    author = {Kumar, Varun Ravi and Yogamani, Senthil and Rashed, Hazem and Sitsu, Ganesh and Witt, Christian and Leang, Isabelle and Milz, Stefan and Mader, Patrick},
    number = {2},
    volume = {6},
    doi = {10.1109/LRA.2021.3062324},
    issn = {23773766}
}

@article{Hasan2022OpticalReview,
    title = {{Optical Camera Communication in Vehicular Applications: A Review}},
    year = {2022},
    journal = {IEEE Transactions on Intelligent Transportation Systems},
    author = {Hasan, Moh Khalid and Ali, Md Osman and Rahman, Md Habibur and Chowdhury, Mostafa Zaman and Jang, Yeong Min},
    number = {7},
    volume = {23},
    doi = {10.1109/TITS.2021.3086409},
    issn = {15580016}
}

@article{Hsiao2018PreventingIssues,
    title = {{Preventing Emergency Vehicle Crashes: Status and Challenges of Human Factors Issues}},
    year = {2018},
    journal = {Human Factors},
    author = {Hsiao, Hongwei and Chang, Joonho and Simeonov, Peter},
    number = {7},
    pages = {1048--1072},
    volume = {60},
    doi = {10.1177/0018720818786132}
}

@misc{UNECE2016RegulationUNECE,
    title = {{Regulation No 48 of the Economic Commission for Europe of the United Nations (UNECE)}},
    year = {2016},
    booktitle = {Uniform provisions concerning the approval of vehicles with regard to the installation of lighting and light-signalling devices [2016/1723]},
    author = {{UNECE}},
    pages = {125--242}
}

@article{VishwasVenkat2023ReviewVehicles,
    title = {{Review and analysis of the properties of 360-degree surround view cameras in autonomous vehicles}},
    year = {2023},
    journal = {International Journal of Science and Research Archive},
    author = {{Vishwas Venkat} and {Raja Reddy}},
    number = {1},
    volume = {8},
    doi = {10.30574/ijsra.2023.8.2.0333}
}

@article{Zhang2024RS-DETR:RT-DETR,
    title = {{RS-DETR: An Improved Remote Sensing Object Detection Model Based on RT-DETR}},
    year = {2024},
    journal = {Applied Sciences},
    author = {Zhang, Hao and Ma, Zheng and Li, Xiang},
    number = {22},
    volume = {14},
    doi = {10.3390/app142210331},
    issn = {2076-3417}
}

@article{Mecocci2024RTAIAED:YOLOv8,
    title = {{RTAIAED: A Real-Time Ambulance in an Emergency Detector with a Pyramidal Part-Based Model Composed of MFCCs and YOLOv8}},
    year = {2024},
    journal = {Sensors},
    author = {Mecocci, Alessandro and Grassi, Claudio},
    number = {7},
    volume = {24},
    url = {https://www.mdpi.com/1424-8220/24/7/2321},
    doi = {10.3390/s24072321},
    issn = {1424-8220}
}

@article{Yatbaz2024Run-TimeRepresentations,
    title = {{Run-Time Introspection of 2D Object Detection in Automated Driving Systems Using Learning Representations}},
    year = {2024},
    journal = {IEEE Transactions on Intelligent Vehicles},
    author = {Yatbaz, Hakan Yekta and Dianati, Mehrdad and Koufos, Konstantinos and Woodman, Roger},
    number = {6},
    pages = {5033--5046},
    volume = {9},
    doi = {10.1109/TIV.2024.3385531}
}

@misc{Badue2021Self-drivingSurvey,
    title = {{Self-driving cars: A survey}},
    year = {2021},
    booktitle = {Expert Systems with Applications},
    author = {Badue, Claudine and Guidolini, Rânik and Carneiro, Raphael Vivacqua and Azevedo, Pedro and Cardoso, Vinicius B. and Forechi, Avelino and Jesus, Luan and Berriel, Rodrigo and Paix{\~{a}}o, Thiago M. and Mutz, Filipe and de Paula Veronese, Lucas and Oliveira-Santos, Thiago and De Souza, Alberto F.},
    volume = {165},
    doi = {10.1016/j.eswa.2020.113816},
    issn = {09574174}
}

@article{ChehidaDouss2023State-of-the-artVulnerabilities,
    title = {{State-of-the-art survey of in-vehicle protocols and automotive Ethernet security and vulnerabilities}},
    year = {2023},
    journal = {Mathematical Biosciences and Engineering},
    author = {Chehida Douss, Aida Ben and Abassi, Ryma and Sauveron, Damien},
    number = {9},
    volume = {20},
    doi = {10.3934/mbe.2023761},
    issn = {15510018}
}

@article{Kumar2023Surround-ViewChallenges,
    title = {{Surround-View Fisheye Camera Perception for Automated Driving: Overview, Survey {\&} Challenges}},
    year = {2023},
    journal = {IEEE Transactions on Intelligent Transportation Systems},
    author = {Kumar, Varun Ravi and Eising, Ciaran and Witt, Christian and Yogamani, Senthil Kumar},
    number = {4},
    volume = {24},
    doi = {10.1109/TITS.2023.3235057},
    issn = {15580016}
}

@misc{Gonzalez-Saavedra2022SurveyVision,
    title = {{Survey of Cooperative Advanced Driver Assistance Systems: From a Holistic and Systemic Vision}},
    year = {2022},
    booktitle = {Sensors},
    author = {Gonz{\'{a}}lez-Saavedra, Juan Felipe and Figueroa, Miguel and C{\'{e}}spedes, Sandra and Montejo-S{\'{a}}nchez, Samuel},
    number = {8},
    volume = {22},
    doi = {10.3390/s22083040},
    issn = {14248220}
}

@inproceedings{Dave2020TAO:Object,
    title = {{TAO: A Large-Scale Benchmark for Tracking Any Object}},
    year = {2020},
    booktitle = {Lecture Notes in Computer Science (including subseries Lecture Notes in Artificial Intelligence and Lecture Notes in Bioinformatics)},
    author = {Dave, Achal and Khurana, Tarasha and Tokmakov, Pavel and Schmid, Cordelia and Ramanan, Deva},
    volume = {12350 LNCS},
    doi = {10.1007/978-3-030-58558-7{\_}26},
    issn = {16113349}
}

@misc{SAEInternationalRecommendedPractice2021TaxonomyVehicles,
    title = {{Taxonomy and Definitions for Terms Related to Driving Automation Systems for On-Road Motor Vehicles}},
    year = {2021},
    author = {{SAE International Recommended Practice}},
    number = {J3016{\_}202104},
    month = {4},
    doi = {10.4271/J3016{\_}202104}
}

@inproceedings{Cordts2016TheUnderstanding,
    title = {{The Cityscapes Dataset for Semantic Urban Scene Understanding}},
    year = {2016},
    booktitle = {Proceedings of the IEEE Computer Society Conference on Computer Vision and Pattern Recognition},
    author = {Cordts, Marius and Omran, Mohamed and Ramos, Sebastian and Rehfeld, Timo and Enzweiler, Markus and Benenson, Rodrigo and Franke, Uwe and Roth, Stefan and Schiele, Bernt},
    volume = {2016-December},
    doi = {10.1109/CVPR.2016.350},
    issn = {10636919}
}

@article{Everingham2010TheChallenge,
    title = {{The Pascal Visual Object Classes (VOC) Challenge}},
    year = {2010},
    journal = {International Journal of Computer Vision},
    author = {Everingham, Mark and Van Gool, Luc and Williams, Christopher K I and Winn, John and Zisserman, Andrew},
    number = {2},
    pages = {303--338},
    volume = {88},
    publisher = {Springer},
    url = {https://doi.org/10.1007/s11263-009-0275-4},
    doi = {10.1007/s11263-009-0275-4}
}

@article{Yao2023TL-Detector:Vehicles,
    title = {{TL-Detector: Lightweight Based Real-Time Traffic Light Detection Model for Intelligent Vehicles}},
    year = {2023},
    journal = {IEEE Transactions on Intelligent Transportation Systems},
    author = {Yao, Zikai and Liu, Qiang and Xie, Qian and Li, Qing},
    number = {9},
    volume = {24},
    doi = {10.1109/TITS.2023.3267430},
    issn = {15580016}
}

@article{Lin2024TrafficIdentification,
    title = {{Traffic Light Detection and Recognition Using a Two-Stage Framework From Individual Signal Bulb Identification}},
    year = {2024},
    journal = {IEEE Access},
    author = {Lin, Huei-Yung and Lin, Ssu-Yun and Tu, Kai-Chun},
    number = {},
    pages = {132279--132289},
    volume = {12},
    doi = {10.1109/ACCESS.2024.3446277}
}

@article{Lisov2023UsingDetection,
    title = {{Using convolutional neural networks for acoustic-based emergency vehicle detection}},
    year = {2023},
    journal = {Modern Transportation Systems and Technologies},
    author = {Lisov, Andrey A. and Kulganatov, Askar Z. and Panishev, Sergei A.},
    number = {1},
    volume = {9},
    doi = {10.17816/transsyst20239195-107}
}

@article{Lu2023VehicleChallenges,
    title = {{Vehicle Computing: Vision and challenges}},
    year = {2023},
    journal = {Journal of Information and Intelligence},
    author = {Lu, Sidi and Shi, Weisong},
    number = {1},
    volume = {1},
    doi = {10.1016/j.jiixd.2022.10.001},
    issn = {29497159}
}

@article{Geiger2013VisionDataset,
    title = {{Vision meets robotics: The KITTI dataset}},
    year = {2013},
    journal = {International Journal of Robotics Research},
    author = {Geiger, A. and Lenz, P. and Stiller, C. and Urtasun, R.},
    number = {11},
    volume = {32},
    doi = {10.1177/0278364913491297},
    issn = {02783649}
}

@inproceedings{Horgan2015Vision-BasedAdvances,
    title = {{Vision-Based Driver Assistance Systems: Survey, Taxonomy and Advances}},
    year = {2015},
    booktitle = {IEEE Conference on Intelligent Transportation Systems, Proceedings, ITSC},
    author = {Horgan, Jonathan and Hughes, Ciaran and McDonald, John and Yogamani, Senthil},
    volume = {2015-October},
    doi = {10.1109/ITSC.2015.329}
}

@article{Wang2023VV-YOLO:YOLOv4,
    title = {{VV-YOLO: A Vehicle View Object Detection Model Based on Improved YOLOv4}},
    year = {2023},
    journal = {Sensors},
    author = {Wang, Yinan and Guan, Yingzhou and Liu, Hanxu and Jin, Lisheng and Li, Xinwei and Guo, Baicang and Zhang, Zhe},
    number = {7},
    volume = {23},
    doi = {10.3390/s23073385},
    issn = {14248220}
}

@inproceedings{Yogamani2019WoodScape:Driving,
    title = {{WoodScape: A multi-task, multi-camera fisheye dataset for autonomous driving}},
    year = {2019},
    booktitle = {Proceedings of the IEEE International Conference on Computer Vision},
    author = {Yogamani, Senthil and Witt, Christian and Rashed, Hazem and Nayak, Sanjaya and Mansoor, Saquib and Varley, Padraig and Perrotton, Xavier and Odea, Derek and Perez, Patrick and Hughes, Ciaran and Horgan, Jonathan and Sistu, Ganesh and Chennupati, Sumanth and Uricar, Michal and Milz, Stefan and Simon, Martin and Amende, Karl},
    volume = {2019-October},
    doi = {10.1109/ICCV.2019.00940},
    issn = {15505499}
}

@misc{Wang2024YOLOv10:Detection,
    title = {{YOLOv10: Real-Time End-to-End Object Detection}},
    year = {2024},
    author = {Wang, Ao and Chen, Hui and Liu, Lihao and Chen, Kai and Lin, Zijia and Han, Jungong and Ding, Guiguang},
    arxivId = {2405.14458}
}

@misc{Hussain2024YOLOv5Vision,
    title = {{YOLOv5, YOLOv8 and YOLOv10: The Go-To Detectors for Real-time Vision}},
    year = {2024},
    author = {Hussain, Muhammad},
    arxivId = {2407.02988}
}

@inproceedings{Real2017YouTube-BoundingBoxes:Video,
    title = {{YouTube-BoundingBoxes: A large high-precision human-annotated data set for object detection in video}},
    year = {2017},
    booktitle = {Proceedings - 30th IEEE Conference on Computer Vision and Pattern Recognition, CVPR 2017},
    author = {Real, Esteban and Shlens, Jonathon and Mazzocchi, Stefano and Pan, Xin and Vanhoucke, Vincent},
    volume = {2017-January},
    doi = {10.1109/CVPR.2017.789}
}

\section{Biography Section}
 

\begin{IEEEbiographynophoto}{Francisco Vacalebri-Lloret}
was born in 1998 in La Vila Joiosa, Alacant, Spain, he earned his Bachelor's degree in Sound and Image Engineering in Telecommunications. He subsequently completed a Master’s degree in Telecommunications Engineering as well as a Master’s degree in Electronic Systems. With experience in the robotics industry, his research is focused on computer vision and artificial intelligence applied to visual systems and multimodal analysis. His work emphasizes the development and integration of advanced algorithms, sensor fusion techniques, and robust system architectures to address complex challenges in autonomous and intelligent systems.
\end{IEEEbiographynophoto}
\begin{IEEEbiographynophoto}{Lucas Banchero} 
was born in Uruguay. He received his degree in Telecommunications Engineering with a specialization in Sound and Image from the University of Málaga, where he also completed a master's degree in Acoustic Engineering. Since 2023, he has been pursuing a Ph.D. in Telecommunications at the Polytechnic University of Valencia. His research focuses on the industrial applications of sound, leveraging artificial intelligence for tasks such as acoustic scene analysis, signal processing, and sound-based event detection. His work spans various sectors, including the automotive industry, where he contributes to the development of Advanced Driver Assistance Systems (ADAS), and heavy industry, where he explores intelligent sound protection systems to enhance worker safety.
\end{IEEEbiographynophoto}
\begin{IEEEbiographynophoto}{Jose J. Lopez} (Member, IEEE) was born in Valencia, Spain in 1969. He received M.S. and Ph.D. degrees in Telecommunications Engineering from Universitat Politècnica de València (UPV) in 1992 and 1999, respectively. He joined the teaching career in 1993 as assistant lecturer and has progressed through various teaching positions until becoming Full Professor in February 2011, a position he continues to hold.
Dr. Lopez has supervised 10 doctoral theses and has served as Deputy Director of Innovation and Technology Transfer at the Institute of Telecommunications and Multimedia Applications (ITEAM), the most prestigious research institute at UPV. Throughout his career, he has led as Principal Investigator more than 40 R\&D projects. He has authored or co-authored more than 60 scientific papers in prestigious journals of signal processing and acoustics, and more than 160 papers in conference proceedings. His research interests encompass audio, acoustics, and multimedia, with current focus on digital audio processing in spatial sound, sound perception, signal processing algorithms, real-time multimedia systems development, and machine learning and artificial intelligence applications to sound and image. Dr. Lopez currently serves as vice-chair of the Spanish Section of the Audio Engineering Society.
\end{IEEEbiographynophoto}
\begin{IEEEbiographynophoto}{Jose M. Mossi} 
received his Ingeniero de Telecomunicación degree from Universidad Politécnica de Madrid  in 1991, and his Ph.D. from Universitat Politècnica de València (UPV) in 1998.  Since 1991, he has been a member of the Department of Communication at UPV where, from 1999,  he is  Profesor Titular
He has participated in numerous national and European research projects and has published over 40 papers in international technical journals and at renowned conferences in the fields of transportation, medical, and sports image processing.
\end{IEEEbiographynophoto}
\end{document}